\documentclass{article} %
\usepackage{preprint,times}

\usepackage{amsmath,amsfonts,bm}

\def\eqref#1{equation~\ref{#1}}

\def\1{\bm{1}}

\DeclareMathAlphabet{\mathsfit}{\encodingdefault}{\sfdefault}{m}{sl}
\SetMathAlphabet{\mathsfit}{bold}{\encodingdefault}{\sfdefault}{bx}{n}

\usepackage{hyperref}
\usepackage{url}
\usepackage{graphicx}
\usepackage{xcolor}
\usepackage{subcaption}
\graphicspath{{img/}}
\usepackage{wrapfig}
\usepackage{booktabs, tabularx}
\usepackage{siunitx}
\usepackage{enumitem}

\usepackage{float}
\usepackage{algorithm}
\usepackage[noend]{algpseudocode}
\algrenewcommand{\algorithmiccomment}[1]{\hfill{\color{gray}$\triangleright$ #1}}
\algnewcommand{\Hyperparameters}{\item[\textbf{Hyperparameters:}]}

\title{\raggedright {{B}}idirectional {{V}}oronoi-biased {{E}}xploration Curriculum for
{R}einforcement Learning}

\author{Juri Pfammatter$^{1}$, Kaixian Qu$^{1}$, Clemens Schwarke$^{1,2}$, Victor Klemm$^{1}$ \& Marco Hutter$^{1}$ \\
$^{1}$Robotic Systems Lab, ETH Z\"urich, Z\"urich, Switzerland \\
$^{2}$NVIDIA \\
\texttt{\{jpfammatter, kaixqu, cschwarke, vklemm, mahutter\}@ethz.com} \\
}

\begin{document}

\maketitle

\begin{figure}[H]
    \centering
    \includegraphics[width=0.99\linewidth]{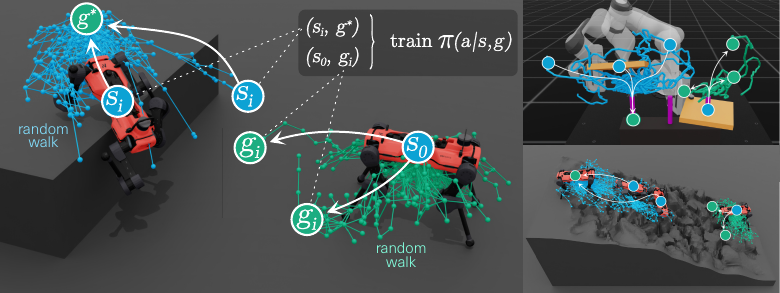}
    \caption{\textbf{BVER grows the curriculum from both ends of the task.}
        Intermediate starts $s_i$ (blue) grow from the target goal $g^\star$ and intermediate
        goals $g_i$ (green) from the initial distribution $s_0 \sim \rho_0$; one policy trains on both
        (white arrows) with sparse rewards and no demonstrations. Shown: \textit{box climbing} (left),
        \textit{ring-on-peg transfer} (top right), and a scanned terrain (bottom right).}
    \label{fig:teaser}
\end{figure}

\begin{abstract}
    Long-horizon tasks with sparse rewards pose an exploration bottleneck for goal-conditioned reinforcement learning: a policy started from the initial state rarely reaches the goal and receives no learning signal. Reference motions, hand-designed curricula, and shaped rewards supply this signal but require demonstrations or task-specific engineering; automatic start-state and goal curricula avoid this but typically expand from one side only, so the full distance to the target must be covered from that side.
    We propose the Bidirectional Voronoi-biased Exploration curriculum for Reinforcement learning (BVER), which expands from both ends at once. Inspired by bidirectional RRT planning, BVER grows start states outward from the goal and goals outward from the initial state distribution, biases both toward unexplored task space, and steers them toward each other, training one goal-conditioned policy on both.
    On point-mass mazes, quadrupedal box climbing, and robot-arm ring-on-peg transfer, BVER learns faster than all compared reference-free curricula. On box climbing, it reaches 95\% success on a \SI{0.4}{\meter} box in roughly 65\% fewer iterations than the best of them, is the only one of them to learn to climb a \SI{0.7}{\meter} box, and yields a policy robust to start, goal, and yaw variation. Without a demonstration, it approaches the sample efficiency of reference-based curricula on the \SI{0.4}{\meter} box and on ring-on-peg transfer. Ablations show that expanding from both ends outperforms either direction alone.
    Project website, videos, and code: \url{https://bver-rl.github.io/}
\end{abstract}

\section{Introduction}\label{sec:introduction}

Many robotic tasks are naturally posed as reaching a goal: a quadruped that should climb onto a box is rewarded for standing on top of it, not for how it gets there. With such a sparse reward and a long horizon, a policy started from the initial state distribution rarely reaches the goal and thus receives no learning signal~\citep{florensa2017reverse}: policy gradient methods such as PPO~\citep{schulman2017proximal} learn from the returns of the trajectories the policy currently generates, and a reward that none of them collects cannot guide it.

In robot control, the missing signal is typically supplied by terrain curricula with dense rewards~\citep{rudin2022learning, lee2020learning}, reference motions~\citep{fuchioka2023opt}, hand-designed curricula~\citep{atanassov2024curriculum}, or curiosity rewards~\citep{schwarke2023curiosity}. These have enabled highly capable controllers but require a demonstration, task-specific engineering, or reward terms beyond the task objective.

Automatic curriculum learning replaces this design effort with intermediate tasks generated during training~\citep{narvekar2020curriculum, portelas2020automatic}. Start-state curricula begin closer to the goal, with start states taken from a reference trajectory~\citep{peng2018deepmimic, resnick2018backplay} or from random walks outward from the goal~\citep{florensa2017reverse}, and goal curricula keep the start fixed and propose intermediate goals~\citep{florensa2018automatic, ren2019exploration, cho2023outcome}. Both expand from one end only: the frontier must cover the entire distance to the fixed end, and the policy trains on the target task from that side only.

\begin{figure}[!t]
    \centering
    \includegraphics[width=0.99\linewidth]{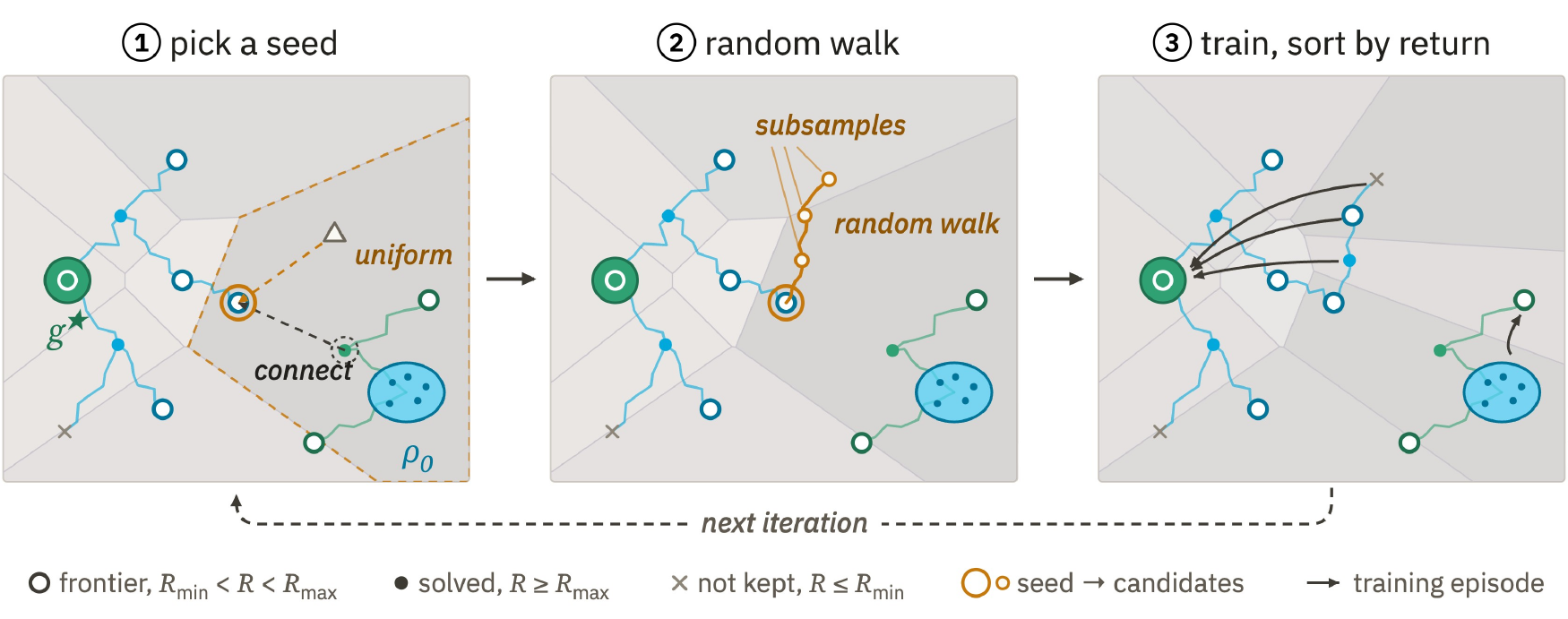}
    \caption{BVER's expansion mechanism. Intermediate starts $s_i$ (blue) grow from the target goal $g^\star$ and goals $g_i$ (green) from the initial distribution $\rho_0$. Random walks start from the states nearest to uniform task-space samples, favoring states whose Voronoi cell (shaded) reaches into unexplored space, or nearest to the other direction's solved states to connect the two (dashed).}\label{fig:mechanism}
\end{figure}

Two mechanisms from sampling-based motion planning address this limitation. Rapidly-exploring random trees~\citep[RRT;][]{lavalle2001randomized} extend the vertex nearest to a uniform sample of the space, so growth is biased toward large unexplored regions (the Voronoi bias), and RRT-Connect~\citep{kuffner2000rrt} grows one tree from the start and one from the goal and extends each toward the other until they meet. BVER (Bidirectional Voronoi-biased Exploration curriculum for Reinforcement learning) transfers both to curriculum generation. It grows start states outward from the goal by random walks (the backwards expansion) and goals outward from the initial state distribution in the same way (the forwards expansion), connects the two sets, and trains a single goal-conditioned policy on both. Like reverse curriculum generation~\citep[RC;][]{florensa2017reverse}, BVER keeps only tasks of intermediate difficulty for the current policy, while its expansion covers the task space (Figure~\ref{fig:mechanism}).

Our contributions are:

\begin{itemize}[leftmargin=16pt, topsep=0.0em]
    \item A bidirectional start--goal curriculum for sparse-reward, long-horizon goal-conditioned RL that generates both intermediate starts and goals, without reference motions, shaped rewards, or a hand-designed curriculum.
    \item An expansion mechanism that carries the Voronoi bias of RRT and the connect step of RRT-Connect over to curriculum generation to propose feasible states as tasks.
    \item Experiments showing that BVER learns fastest among the reference-free curricula on all six compared tasks, is the only one of them to learn to climb the \SI{0.7}{\meter} box, transfers to scanned terrains with the same hyperparameters, and outperforms either expansion direction alone.
\end{itemize}

\section{Related work}\label{sec:related_work}

\paragraph{Automatic curriculum learning:}
Automatic curriculum learning (ACL) adapts the training tasks to the learner~\citep{narvekar2020curriculum, portelas2020automatic}. Environment curricula sample generator parameters by learning progress~\citep{portelas2020teacher}, replay levels by value loss~\citep{jiang2021prioritized}, generate regret-maximizing levels~\citep{dennis2020paired, parkerholder2022accel}, or widen domain randomization as performance improves~\citep{akkaya2019solving}; reward curricula add intrinsic rewards such as RND~\citep{burda2018exploration}, whose authors highlight global exploration as an important direction. BVER is complementary, changing only where episodes start and which goal they pursue.

\paragraph{Start-state curricula:}
Reverse curriculum generation~\citep[RC;][]{florensa2017reverse} grows start states outward from the goal by random-action rollouts and keeps those of intermediate difficulty, using resets but no demonstrations; BaRC~\citep{ivanovic2019barc} uses backward reachable sets of an approximate model instead. Demonstration-based methods reset to states of a reference trajectory~\citep{salimans2018montezuma, nair2018overcoming}, sampled uniformly~\citep[RSI;][]{peng2018deepmimic}, moved backward on a schedule~\citep[Backplay;][]{resnick2018backplay}, or prioritized by success rate~\citep[RFCL;][]{tao2024reverse}. Go-Explore~\citep{ecoffet2021first} and PBCS~\citep{matheron2020pbcs} return to rarely visited states from an archive and explore from there. Similarly, BVER seeds its random walks at stored states.

\paragraph{Goal curricula:}
Goal curricula keep the initial state distribution and propose goals: GoalGAN~\citep{florensa2018automatic} generates goals of intermediate difficulty; HGG~\citep{ren2019exploration} and OUTPACE~\citep{cho2023outcome} select achieved states by progress toward the target, and DISCOVER~\citep{diazbone2025discover} by estimated achievability, novelty, and relevance; Skew-Fit~\citep{pong2020skewfit} and MEGA~\citep{pitis2020mega} select achieved states of low density; and C-Planning~\citep{zhang2022cplanning} commands waypoints between start and goal under a learned distance. VaPRL~\citep{sharma2021vaprl} and IBC~\citep{kim2023ibc} build such curricula from replay-buffer states and, unlike BVER, need no resets. BVER builds on both lines of work, growing start states and goals toward each other and seeding them by Voronoi bias instead of visit counts or density.
\paragraph{Sampling-based planning:}
A rapidly-exploring random tree (RRT)~\citep{lavalle2001randomized} extends the vertex nearest to a random sample; vertices are thus selected in proportion to the area of their Voronoi regions, which biases the tree toward unexplored space~\citep{kuffner2000rrt}. RRT-Connect~\citep{kuffner2000rrt} grows trees from start and goal and connects them. R3L~\citep{morere2020r3l} initializes a policy from RRT solutions, and SoRB~\citep{eysenbach2019sorb} searches a graph over replay-buffer states with a learned value function. \citet{florensa2017reverse} proposed combining start-state and goal generation as future work, citing RRT-Connect as the analogue; BVER follows this suggestion, growing a curriculum rather than a path.

\paragraph{Curricula and references in robot learning:}
Legged robots are mostly trained with dense rewards and terrain curricula~\citep{rudin2022learning, lee2020learning}, with an inner-product reward for parkour~\citep{cheng2024extreme}, or with RSI on a trajectory-optimization reference~\citep{fuchioka2023opt}. Hand-designed curricula start jumps at random heights and upward velocities~\citep{atanassov2024curriculum} or grow obstacle dimensions and navigation distances with performance~\citep{hoeller2024anymal}, and \citet{schwarke2023curiosity} extend RND with a curiosity state for sparse-reward loco-manipulation. These methods yield highly capable controllers; BVER targets settings without such guidance.

\section{Preliminaries}\label{sec:preliminaries}

\subsection{Goal-Conditioned Markov Decision Process}\label{sec:problem_formulation}

We model the task as a goal-augmented Markov decision process (GA-MDP)~\citep{liu2022goal} with state space $\mathcal{S}$, action space $\mathcal{A}$, transition kernel $P$, discount factor $\gamma$, initial state distribution $\rho_0$, and goal space $\mathcal{G}$.
A known mapping $\phi: \mathcal{S} \to \mathcal{G}$ projects a state onto the goal it achieves~\citep{andrychowicz2017hindsight}, and a goal $g$ is reached in state $s$ if $\lVert \phi(s) - g \rVert \le \epsilon$ for a tolerance $\epsilon$.
The reward is sparse and binary: the agent receives $+1$ in every step at which the goal is reached and $0$ otherwise. Episodes have a fixed horizon $H$ and do not terminate when the goal is reached.

The agent acts with a goal-conditioned policy $\pi_\theta(a \mid s, g)$~\citep{schaul2015universal}. Since BVER only chooses the start--goal pairs of the rollouts, it is not tied to a particular learner; we use PPO~\citep{schulman2017proximal}, which scales to thousands of parallel simulated environments~\citep{rudin2022learning}, rather than off-policy learning with hindsight relabeling~\citep{andrychowicz2017hindsight}.
We measure a rollout $\tau = (s_0, \ldots, s_{H-1})$ under goal $g$ by its undiscounted return, normalized by the maximum attainable return $H$, i.e., the fraction of the horizon spent at the goal:
\begin{equation}\label{eq:normalized_return}
    R(\tau, g) = \frac{1}{H} \sum_{t=0}^{H-1} \mathbf{1}\big[\lVert \phi(s_t) - g \rVert \le \epsilon\big] \in [0, 1].
\end{equation}
Since the reward accrues in every step at the goal, $R$ reflects how quickly the policy reaches the goal and how reliably it stays there, rather than binary success.
The \emph{target task} is to reach a single target goal $g^\star \in \mathcal{G}$ from the initial state distribution $\rho_0$, and the learning objective is
\begin{equation}\label{eq:objective}
    J(\pi) = \mathbb{E}_{s_0 \sim \rho_0,\; \tau \sim \pi(\cdot \mid \cdot,\, g^\star)} \big[ R(\tau, g^\star) \big].
\end{equation}
Following \citet{florensa2017reverse}, we assume that the simulator can be reset to arbitrary states and simulated from them under random actions, that $\rho_0$ can be sampled, that a goal state $s^{g} \in \mathcal{S}$ with $\phi(s^{g}) = g^\star$ is known, and that a bounded task space $\mathcal{T}$ containing the target task is known and can be sampled uniformly.

\subsection{Curriculum}\label{sec:curriculum}

Curriculum learning trains on a sequence of tasks of increasing difficulty~\citep{narvekar2020curriculum, portelas2020automatic}. In our setting, a task is a start--goal pair, and a \emph{curriculum} decides in every training iteration which pairs the policy is trained on, based on the experience collected with the current policy~\citep{tao2024reverse}; the intermediate tasks are auxiliary, and the objective remains the target task.
Formally, in iteration $k$ the curriculum replaces the start--goal distribution $\rho_0 \times \delta_{g^\star}$ of Equation~\ref{eq:objective} by a distribution $q_k$ over start--goal pairs, while the transition kernel $P$ and the reward, and thus $R$, remain unchanged:
\begin{equation}\label{eq:curriculum}
    J_k(\pi) = \mathbb{E}_{(s_0, g) \sim q_k,\; \tau \sim \pi(\cdot \mid \cdot,\, g)} \big[ R(\tau, g) \big].
\end{equation}
A \emph{start-state curriculum} keeps the goal fixed at $g^\star$ and proposes intermediate start states, $q_k = \sigma_k \times \delta_{g^\star}$~\citep{florensa2017reverse, tao2024reverse}, whereas a \emph{goal curriculum} keeps the start states drawn from $\rho_0$ and proposes intermediate goals, $q_k = \rho_0 \times \mu_k$~\citep{florensa2018automatic, ren2019exploration}. BVER does both at once with one shared goal-conditioned policy (Section~\ref{sec:method}), mixing the two with a ratio $\alpha \in [0, 1]$:
\begin{equation}\label{eq:bver_mixture}
    q_k = \alpha \, \sigma_k \times \delta_{g^\star} + (1 - \alpha) \, \rho_0 \times \mu_k.
\end{equation}

\section{Method}\label{sec:method}
BVER maintains finite sets of states from which the starts and goals of each training iteration are sampled uniformly at fixed ratios. We first describe the backwards expansion, which extends RC's random walks with task space coverage, then apply it to the forwards expansion and combine the two. Algorithm~\ref{alg:bver_overview} summarizes the method, with its subroutines in Appendix~\ref{sec:appendix_algorithms}; Figure~\ref{fig:method} in the appendix gives an overview.

\subsection{Voronoi-Biased Expansion}

\paragraph{Exploration:}
As in RC~\citep{florensa2017reverse}, training starts from valid states obtained by simulating random actions from the goal state (blue in Figure~\ref{fig:mechanism}). After each training iteration, we sort these states $s_i$ by the normalized return (Equation~\ref{eq:normalized_return}) of their rollouts $\tau_i$ into frontier and solved states,
\begin{equation}\label{eq:frontier_solved}
    F = \{ s_i : R_{\min} < R(\tau_i, g^\star) < R_{\max} \}, \qquad S = \{ s_i : R(\tau_i, g^\star) \geq R_{\max} \},
\end{equation}
and discard states with $R \leq R_{\min}$ as too hard for the current policy. Mixing the two buffers at a fixed ratio yields the \emph{mixed} training set, which thus holds mostly states of intermediate difficulty.

To choose the seeds of the next random walks, we add a criterion that promotes coverage of the task space and thus drives the frontier toward the initial distribution: we (1) sample points uniformly from the task space $\mathcal{T}$, (2) find for each sample its nearest neighbor in the mixed training set under the projection $\phi$, and (3) use these neighbors as seeds. A state is thereby selected with probability proportional to the volume of its Voronoi cell in $\mathcal{T}$ (shaded in Figure~\ref{fig:mechanism}), which is largest for states at the edge of the explored region, whose cells reach far into unexplored space. As with the Voronoi bias of RRT~\citep{lavalle2001randomized}, this steers the expansion outward instead of refining regions the curriculum already covers. Each walk then simulates $T_B$ random actions from its seed, and $N_s$ states subsampled along it become candidates for the next rollouts.

\begin{algorithm}[t!]
    \caption{BVER (subroutines in Appendix~\ref{sec:appendix_algorithms})}\label{alg:bver_overview}
    \small
    \begin{algorithmic}[1]
        \Require{initial distribution $\rho_0$, target goal $g^\star$ with goal state $s^{g}$, task space $\mathcal{T}$, goal-conditioned policy $\pi$, environment split $N_{\text{explore}}, N_{\text{exploit}}$}
        \State{} $\mathcal{I}^{\text{fwd}} \gets \{s_0 \sim \rho_0\}$, \; $\mathcal{I}^{\text{bwd}} \gets \{s^{g}\}$; \; $C^{d}$, $F^{d}, S^{d} \gets \emptyset$ for $d \in \{\text{fwd}, \text{bwd}\}$
        \For{each iteration}
        \For{$(d, \bar{d}) \in \{(\text{fwd}, \text{bwd}), (\text{bwd}, \text{fwd})\}$} \Comment{environments split according to Equation~\ref{eq:bver_mixture}}
        \State{} $C^{d} \gets C^{d} \cup \Call{RandomWalk}{\mathcal{I}^{d}, \mathcal{T}, S^{\bar{d}}}$ \Comment{seeds from Equation~\ref{eq:connect}}
        \State{} $D^{d} \gets \Call{Rollout}{\pi, C^{d}, d, N_{\text{explore}}/2}$ \Comment{explore: $(s_0 \sim \rho_0, \phi(\cdot))$ if fwd, $(\cdot, g^\star)$ if bwd}
        \State{} $D^{d} \gets D^{d} \cup \Call{Rollout}{\pi, \mathcal{I}^{d}, d, N_{\text{exploit}}/2}$ \Comment{exploit}
        \State{} $F^{d}, S^{d} \gets \Call{Threshold}{D^{d}, F^{d}, S^{d}, d}$ \Comment{append; threshold according to Equation~\ref{eq:frontier_solved}}
        \State{} $\mathcal{I}^{d} \gets \Call{Mix}{F^{d}, S^{d}}$ \Comment{replay a fraction $\nu$ of solved states}
        \EndFor{}
        \State{} $\pi \gets \Call{Train}{\pi, D^{\text{fwd}} \cup D^{\text{bwd}}}$ \Comment{one shared policy update}
        \EndFor{}
    \end{algorithmic}
\end{algorithm}

\paragraph{Exploitation:} In parallel, $N_{\text{exploit}}$ of the parallel environments continue training on the mixed set, while $N_{\text{explore}}$ train on the new random-walk states. Training on the reward frontier improves the policy, and replaying solved states mitigates catastrophic forgetting. This step is optional: on tasks without forgetting, we omit it, and then the mixed set only seeds the random walks (Appendix~\ref{sec:appendix_algorithms}). Both steps repeat until the policy solves the task from the initial distribution.

\subsection{Bidirectional Expansion}

\paragraph{Forwards expansion:}
The forwards expansion applies the same mechanism with the roles of start and goal exchanged: the policy is trained on pairs $(s_0 \sim \rho_0,\, g_i)$ with goals $g_i$ from the forwards frontier, rather than $(s_i,\, g^\star)$. The random walks are then seeded at states that can be reached, rather than states from which the goal can be reached. Because the walks must be reset to full simulator states, the forwards buffers store the final state of each rollout, sorted by the return of its commanded goal; a rollout that reached and held its goal ends within $\epsilon$ of it (Appendix~\ref{sec:appendix_algorithms}).

\paragraph{Bidirectional expansion:} Combining the two, we train one shared goal-conditioned policy on both $(s_0 \sim \rho_0,\, g_i)$ and $(s_i,\, g^\star)$ (arrows in Figure~\ref{fig:mechanism}), with the training budget split evenly between the two expansions ($\alpha = \frac{1}{2}$ in Equation~\ref{eq:bver_mixture}). The curriculum thus covers the reachable task space from two directions, and the policy learns to reach the target goal from a wide range of start states and to reach a wide range of goals from the initial distribution.

\paragraph{Connecting the two frontiers:} As in RRT-Connect~\citep{kuffner2000rrt}, which grows two trees toward each other until they connect, we steer the two frontiers together by replacing the samples from $\mathcal{T}$ in step (1) with samples from the other direction's solved set (dashed line in Figure~\ref{fig:mechanism}). When expanding the forwards frontier, for example, we sample start states from which the policy already reaches $g^\star$ and seed the walks at their nearest neighbors in the forwards mixed training set, so the intermediate goals grow toward $g^\star$; symmetrically, the intermediate starts grow toward $\rho_0$. Each expansion thus moves into the region the other has already made solvable instead of spreading uniformly over $\mathcal{T}$. To balance coverage against connection, we draw each sample from the mixture
\begin{equation}\label{eq:connect}
    x_{\text{rand}} \sim p_{\text{connect}} \, \mathcal{U}\Bigl(\phi\Bigl(S^{\bar{d}}\Bigr)\Bigr) + (1 - p_{\text{connect}}) \, \mathcal{U}(\mathcal{T}),
\end{equation}
where $S^{\bar{d}}$ is the other direction's solved set and $p_{\text{connect}}$ the connect ratio.

\section{Experiments}\label{sec:experiments}
\begin{figure}[t!]
    \centering
    \captionsetup[subfigure]{justification=centering}
    \hfill
    \begin{subfigure}[t]{0.112\linewidth}
        \centering
        \includegraphics[width=\linewidth]{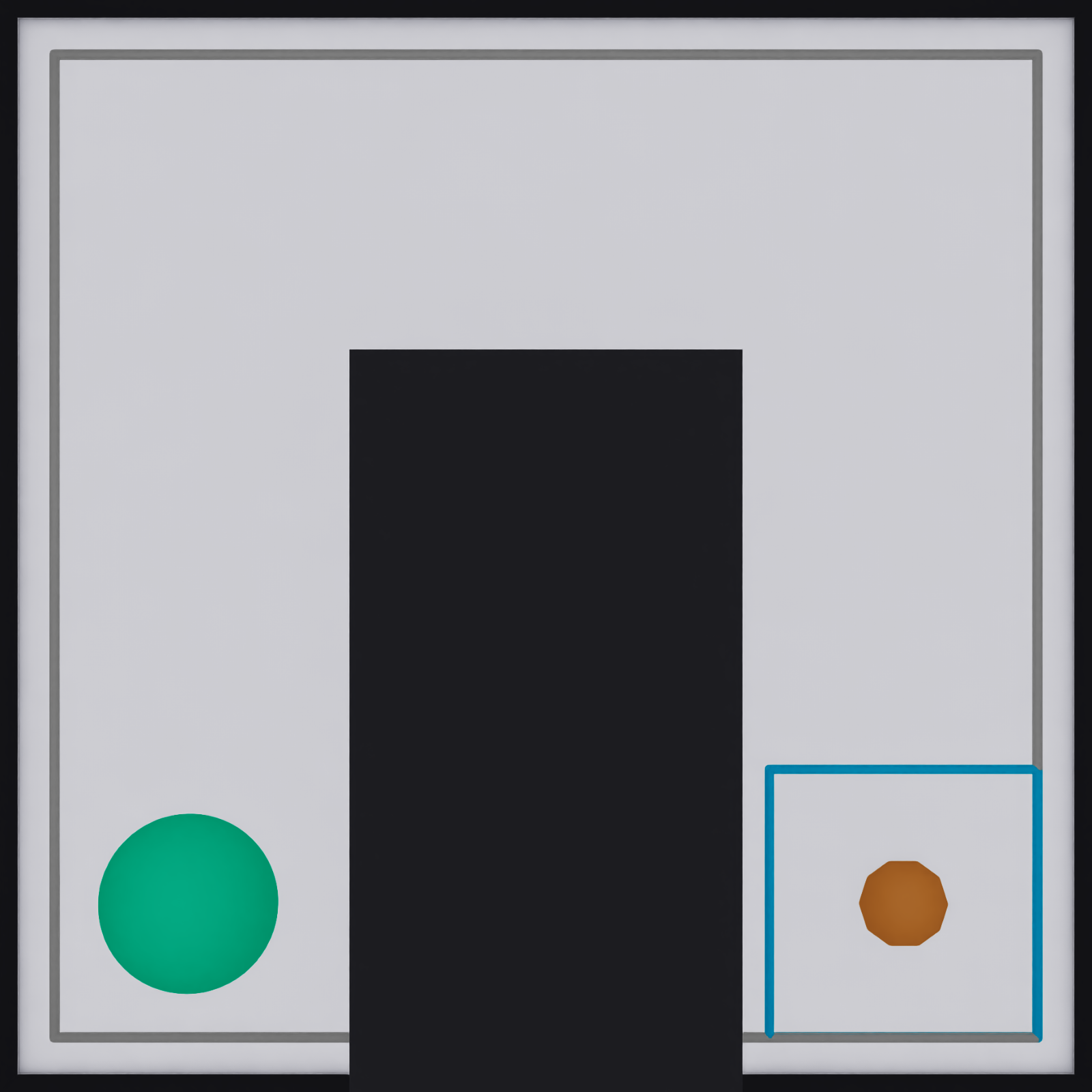}
        \caption{U-maze.}\label{fig:exp_renders_maze_u}
    \end{subfigure}
    \hfill
    \begin{subfigure}[t]{0.112\linewidth}
        \centering
        \includegraphics[width=\linewidth]{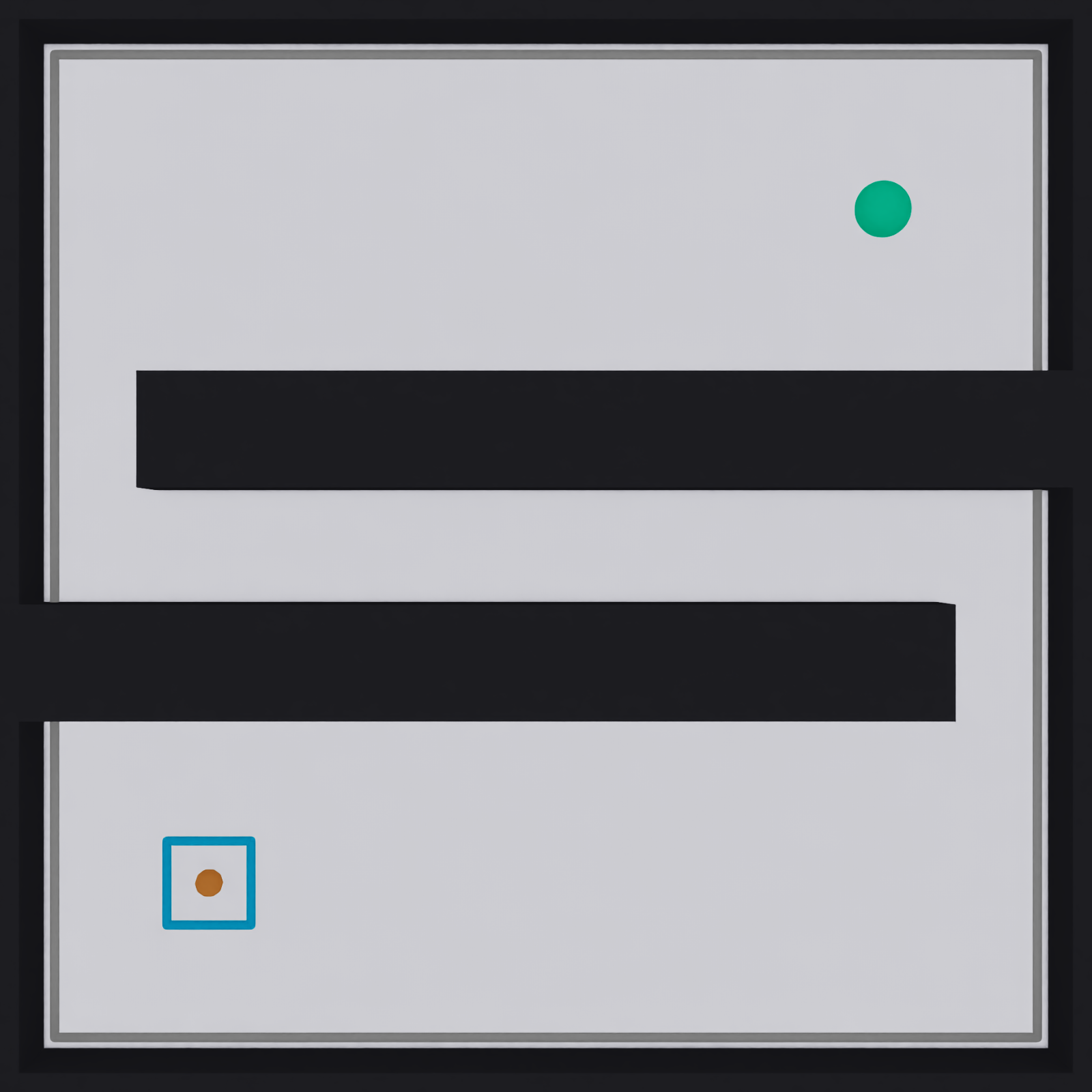}
        \caption{N-maze.}\label{fig:exp_renders_maze_n}
    \end{subfigure}
    \hfill
    \begin{subfigure}[t]{0.112\linewidth}
        \centering
        \includegraphics[width=\linewidth]{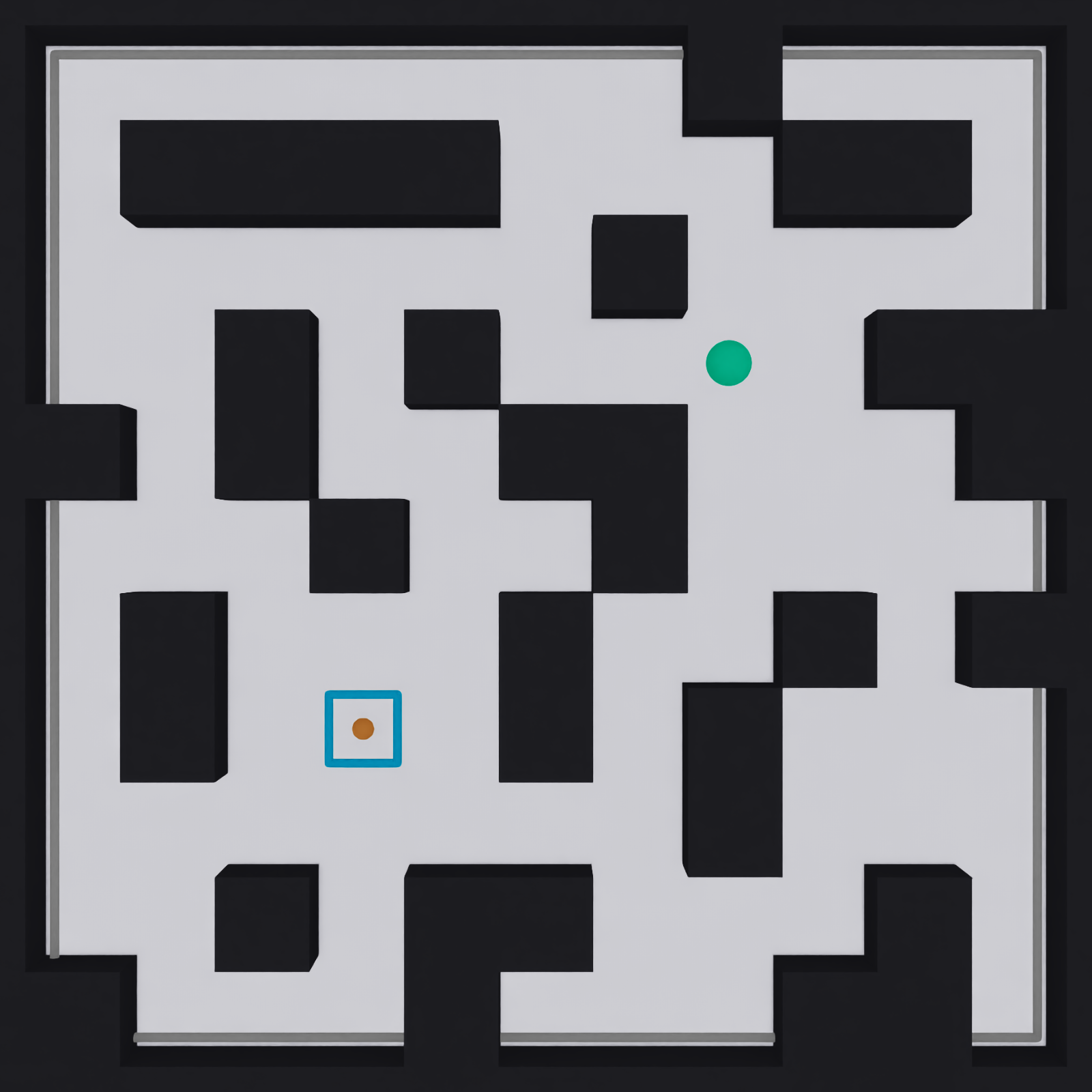}
        \caption{Large maze.}\label{fig:exp_renders_maze_large}
    \end{subfigure}
    \hfill
    \begin{subfigure}[t]{0.2\linewidth}
        \centering
        \includegraphics[width=\linewidth]{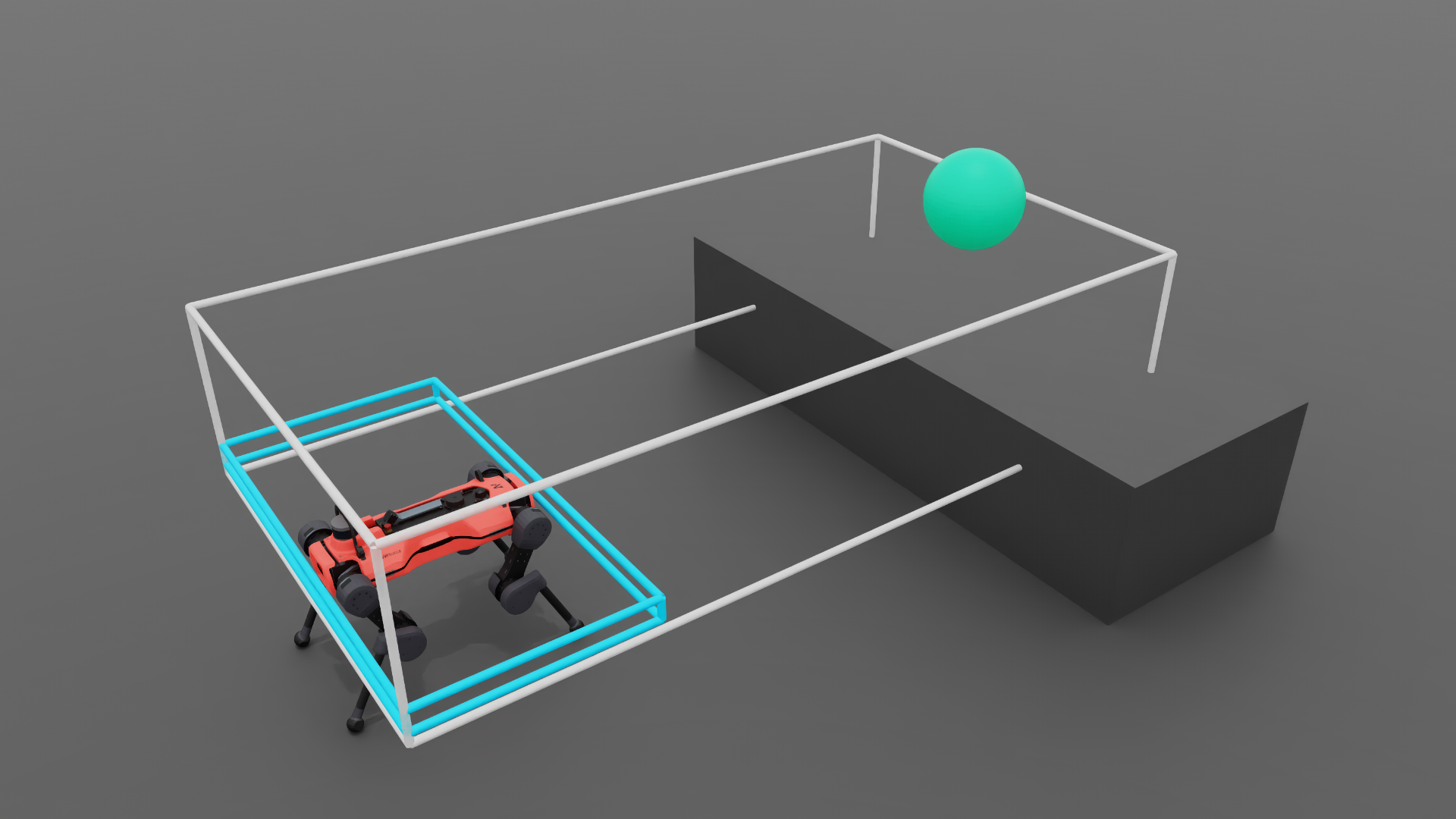}
        \caption{Climb box up.}\label{fig:exp_renders_climb_box_up}
    \end{subfigure}
    \hfill
    \begin{subfigure}[t]{0.2\linewidth}
        \centering
        \includegraphics[width=\linewidth]{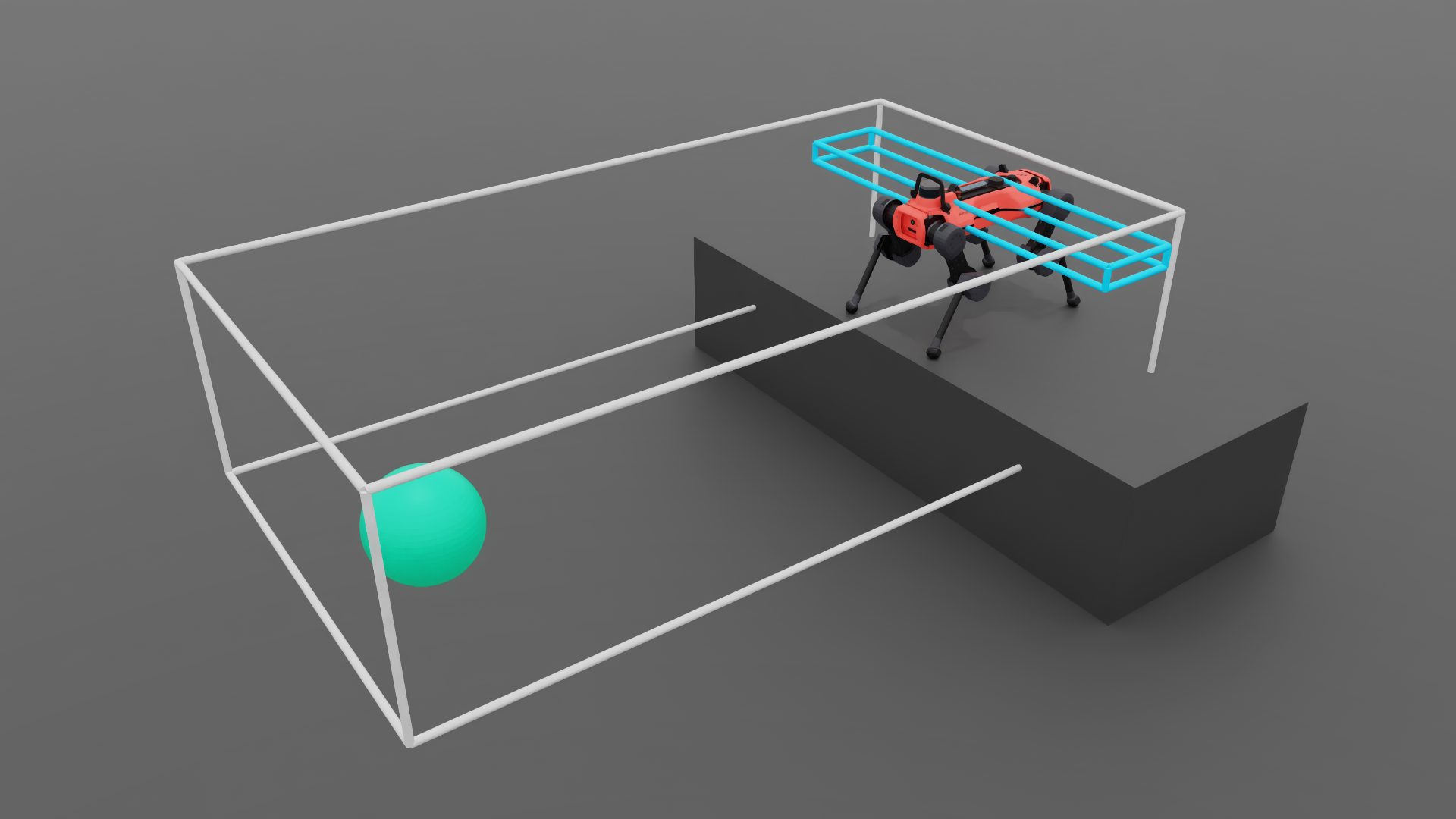}
        \caption{Climb box down.}\label{fig:exp_renders_climb_box_down}
    \end{subfigure}
    \hfill
    \begin{subfigure}[t]{0.2\linewidth}
        \centering
        \includegraphics[width=\linewidth]{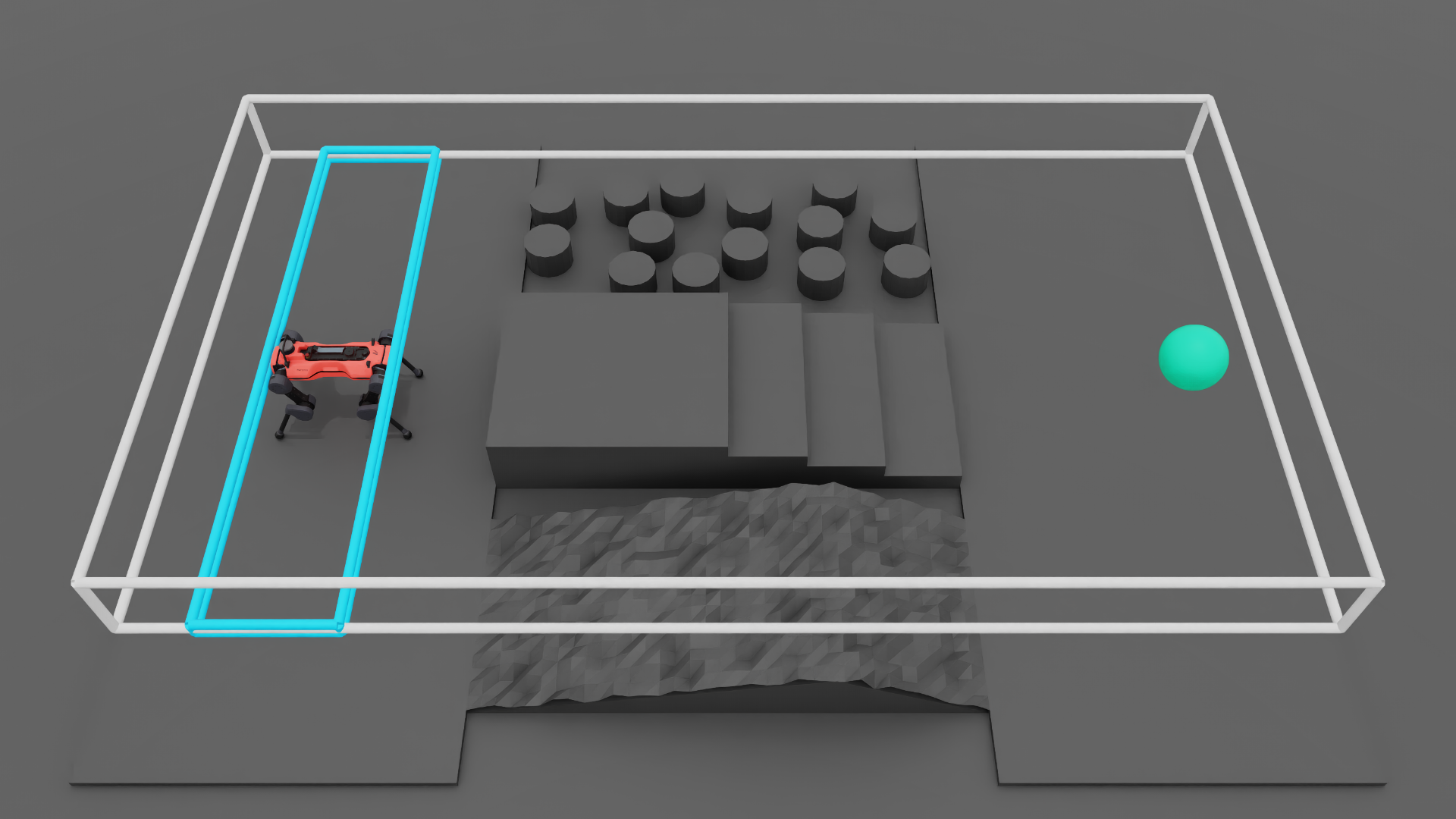}
        \caption{Three-route terrain.}\label{fig:exp_renders_dtsg}
    \end{subfigure}
    \hfill\strut{}\\
    \hfill
    \begin{subfigure}[t]{0.19\linewidth}
        \centering
        \includegraphics[width=\linewidth]{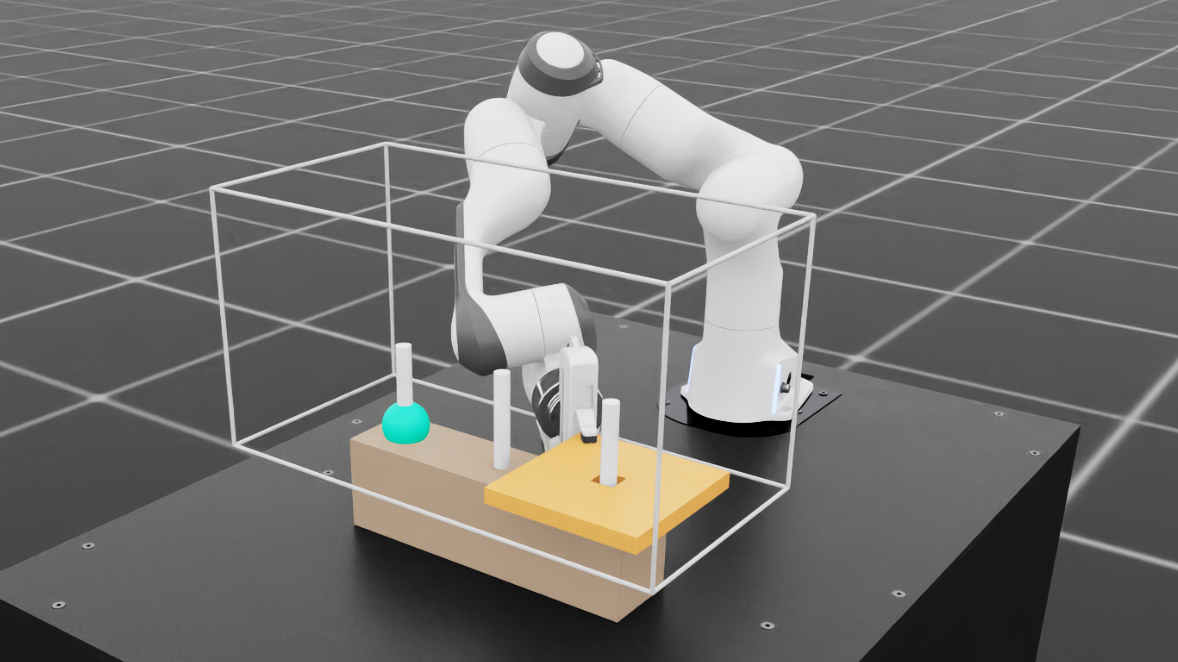}
        \caption{Ring-on-peg transfer.}\label{fig:exp_renders_hanoi_p12}
    \end{subfigure}
    \hfill
    \begin{subfigure}[t]{0.19\linewidth}
        \centering
        \includegraphics[width=\linewidth]{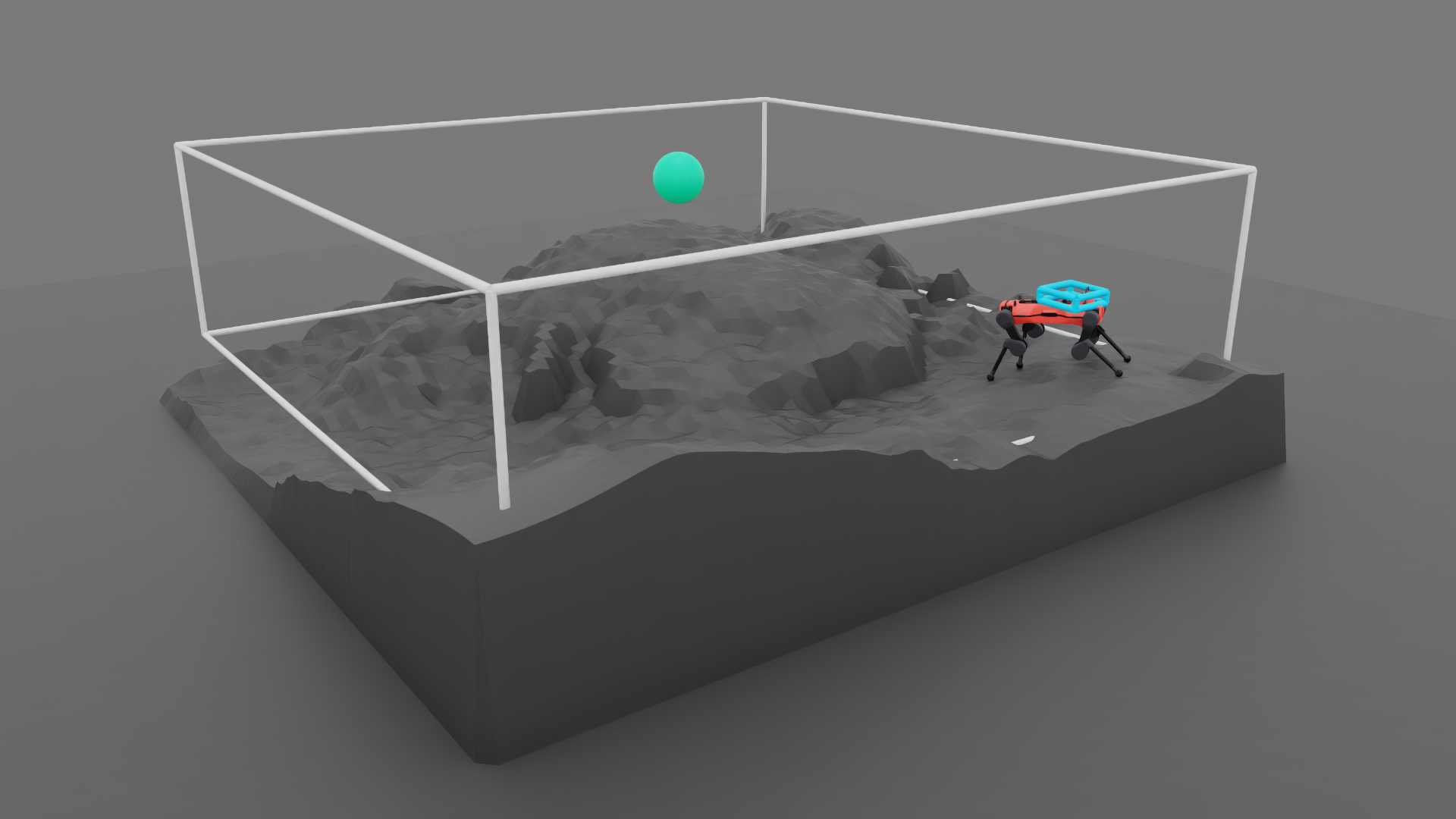}
        \caption{Freestanding boulder I.}\label{fig:exp_renders_freestanding_boulder_I}
    \end{subfigure}
    \hfill
    \begin{subfigure}[t]{0.19\linewidth}
        \centering
        \includegraphics[width=\linewidth]{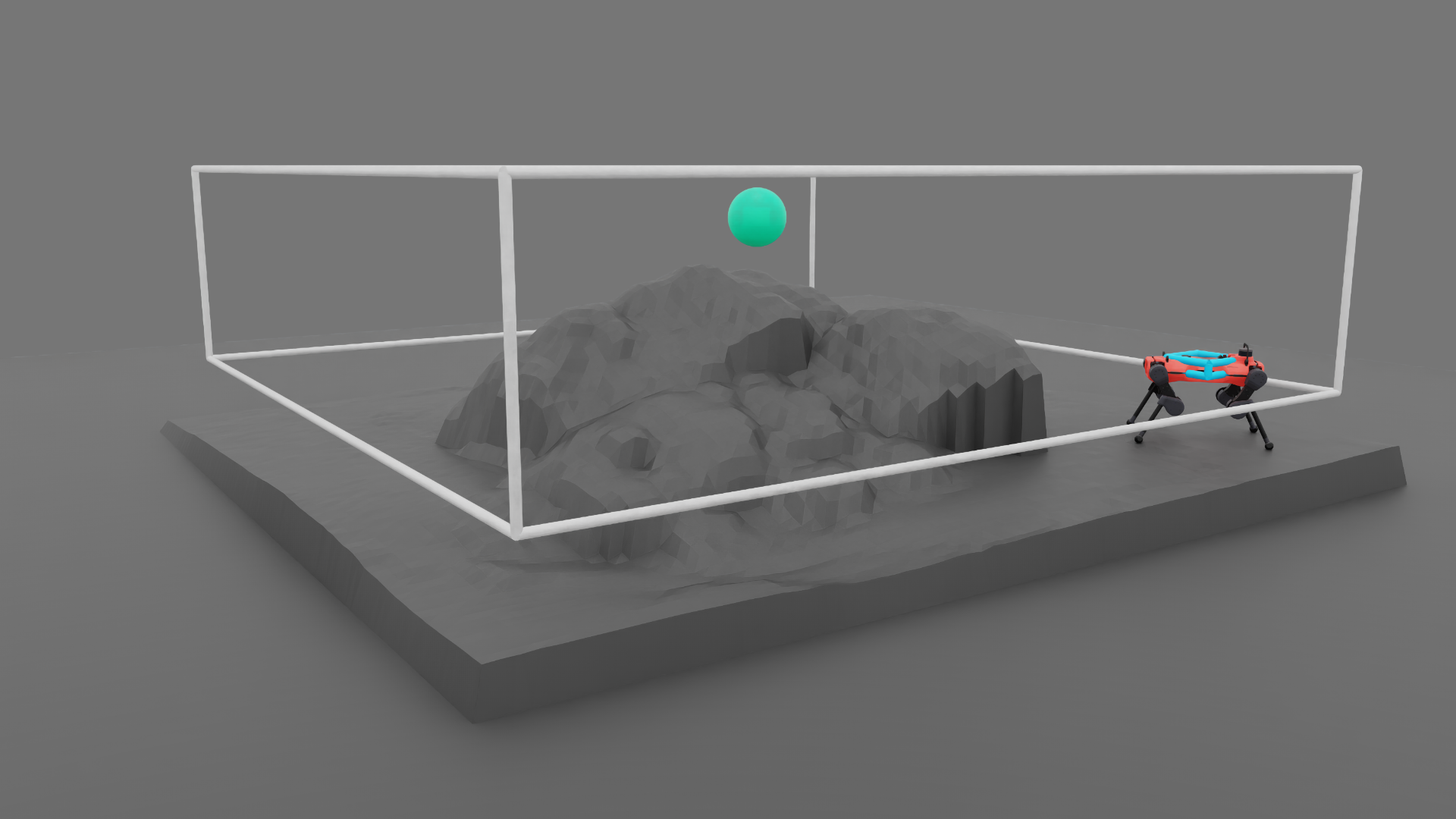}
        \caption{Freestanding boulder II.}\label{fig:exp_renders_freestanding_boulder_II}
    \end{subfigure}
    \hfill
    \begin{subfigure}[t]{0.19\linewidth}
        \centering
        \includegraphics[width=\linewidth]{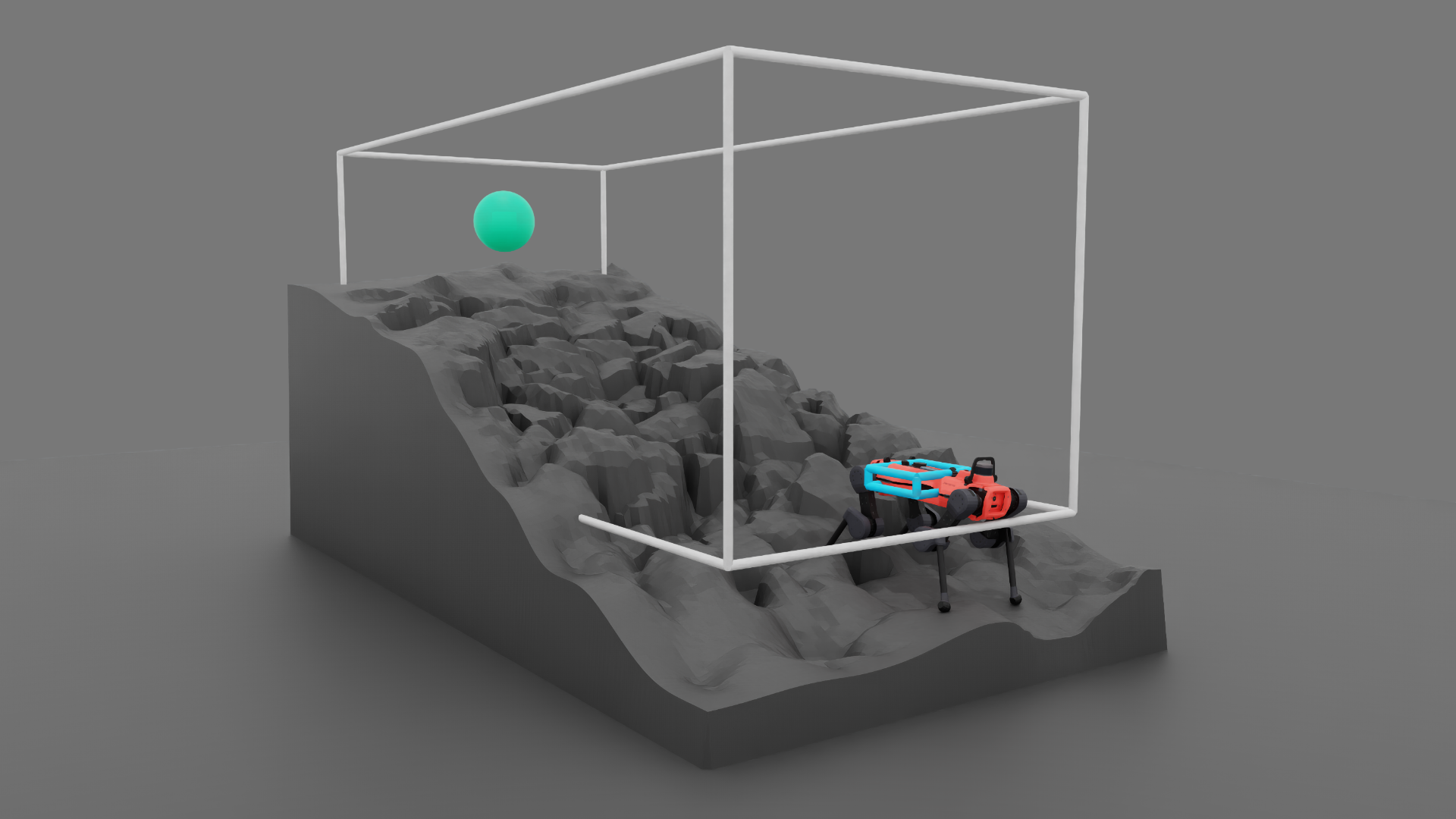}
        \caption{Rocky ascent.}\label{fig:exp_renders_rocky_ascent}
    \end{subfigure}
    \hfill
    \begin{subfigure}[t]{0.19\linewidth}
        \centering
        \includegraphics[width=\linewidth]{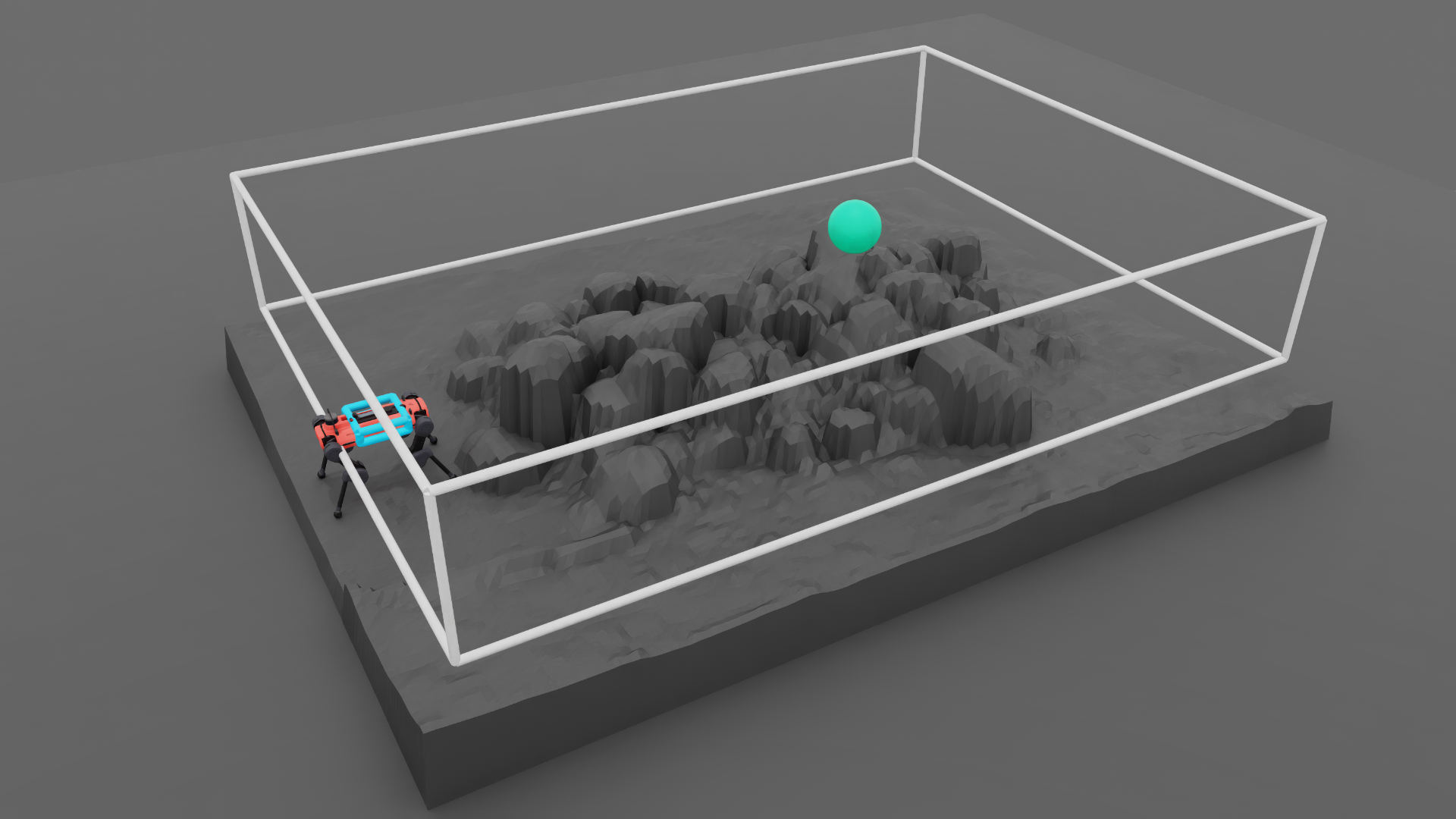}
        \caption{Small stone field.}\label{fig:exp_renders_small_rocks}
    \end{subfigure}
    \hfill
    \caption{Task overview: three \textit{point-mass mazes} (a--c), quadrupedal \textit{box climbing} (d, e), the \textit{three-route terrain} (f), \textit{ring-on-peg transfer} (g), and four scanned terrains (h--k). Cyan indicates the initial distribution, green the target goal, and the white box represents the task space.}\label{fig:experiment_renders}
\end{figure}

We study BVER's sample efficiency and robustness compared to existing curricula (Section~\ref{sec:exp_main}), whether its training covers the whole task space (Section~\ref{sec:exp_emergent}), how it transfers to terrains scanned from real environments (Section~\ref{sec:exp_real_world}), and the contribution of its expansion directions and connect step (Section~\ref{sec:exp_ablations}).
All experiments are implemented in Isaac Lab~\citep{nvidia2025isaaclabgpuacceleratedsimulation} and trained with PPO~\citep{schulman2017proximal}; Figure~\ref{fig:experiment_renders} shows the simulated tasks. Unless stated otherwise, the reward is sparse, without shaping terms. The task space $\mathcal{T}$ is a box in position space, i.e., $\phi(s)$ projects the point-mass, base, or ring position to $(x,y)$ or $(x,y,z)$, which makes goals easy to specify and gives a Euclidean metric for the nearest-neighbor search. Observation and action spaces, rewards, and hyperparameters are detailed in Appendix~\ref{sec:appendix_obs_act}. Learning curves show the mean success rate over five seeds (standard error shaded) against the training iteration; all methods use the same number of parallel environments and simulation steps per iteration. The demonstrations for the reference-based baselines are rollouts of policies trained with BVER.

\subsection{Sample Efficiency and Robustness across Domains}\label{sec:exp_main}
We first test whether BVER improves sample efficiency and robustness over curriculum baselines, and whether these gains hold across embodiments. In the \textit{point-mass mazes} (Figures~\ref{fig:exp_renders_maze_u}--\ref{fig:exp_renders_maze_large}), similar to those used in prior work~\citep{florensa2017reverse,cho2023outcome}, a velocity-controlled point mass must find a path from the initial distribution to the goal. In \textit{box climbing} (Figures~\ref{fig:exp_renders_climb_box_up} and~\ref{fig:exp_renders_climb_box_down}), the quadruped must reach a goal on top of a box from an initial distribution in front of it or, in the symmetric climb-down task, a goal on the ground from the top of the box. In \textit{ring-on-peg transfer} (Figure~\ref{fig:exp_renders_hanoi_p12}), a Franka Panda arm holding a rigidly attached ring must move it from the bottom of one peg to the bottom of another, a reaching task constrained by the pegs. On all tasks, we compare against PPO without a curriculum (\textit{vanilla PPO}), a simple curriculum that samples intermediate starts and goals uniformly from the free task space (\textit{random curriculum}), and the reference-free reverse curriculum (RC)~\citep{florensa2017reverse}. On \textit{ring-on-peg transfer} and \textit{box climbing}, we additionally compare against two reference-based methods that require a demonstration (marked with $\star$): reference state initialization (RSI)~\citep{peng2018deepmimic} and Backplay~\citep{resnick2018backplay}.

\begin{figure}[t!]
    \centering
    \includegraphics[width=\linewidth]{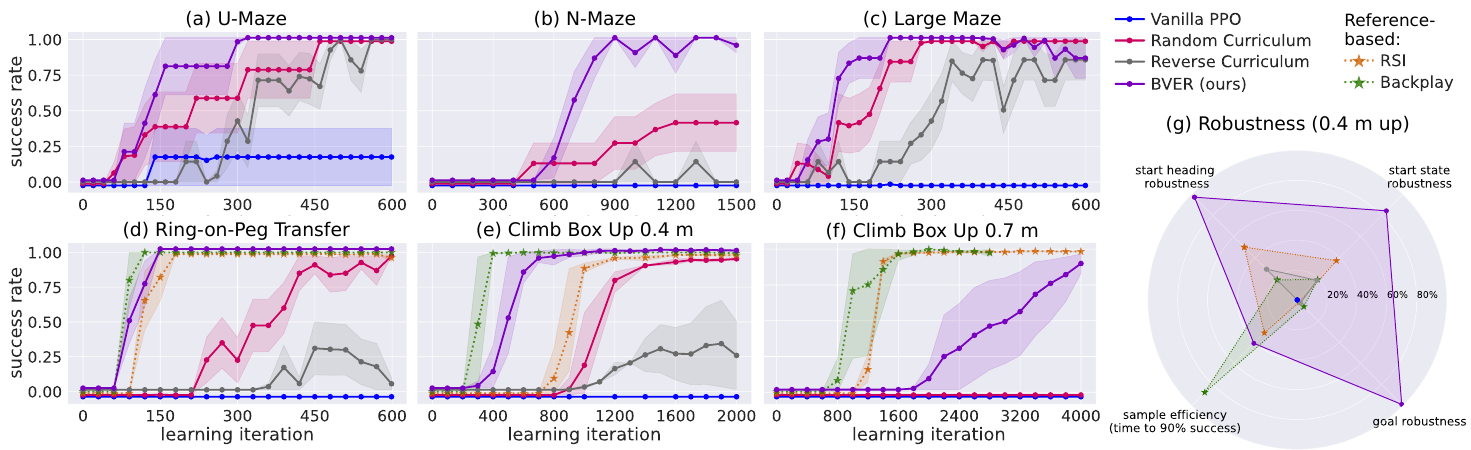}
    \caption{Sample efficiency across domains (a--f; mean over five seeds, standard error shaded) and robustness on \textit{box climbing} (g).
        BVER learns fastest among the reference-free curricula on all tasks, and on \textit{ring-on-peg transfer} and the \SI{0.4}{\meter} box even faster than the reference-based RSI (marked with $\star$). Panel (g) uses the same colors as (a--f).
    }\label{fig:bench_domains}
\end{figure}

\paragraph{Sample Efficiency:}
Figure~\ref{fig:bench_domains} shows the success rate over training. Vanilla PPO succeeds only for a few seeds in the simplest \textit{U-maze} and never on any other task. Initializing the agent closer to the goal makes the reward attainable, but not on every task: the random curriculum plateaus at about 40\% success in the \textit{N-maze}, likely because few of its uniform samples fall into the narrow passages; RC stays below about 35\% in the \textit{N-maze}, on \textit{ring-on-peg transfer}, and on the \SI{0.4}{\meter} box, as does BVER's backwards expansion alone on the first two (Table~\ref{tab:ablation_table}); and neither succeeds on the \SI{0.7}{\meter} box. We attribute RC's low success rate to the lack of an explicit incentive to expand toward $\rho_0$: on \textit{ring-on-peg transfer}, for example, we observed that its start states spread into the free space above the pegs rather than toward the starting peg.

BVER learns fastest among the reference-free curricula on all tasks. On \textit{ring-on-peg transfer}, it exceeds 95\% success in about 150 iterations, four times fewer than the random curriculum, and lies between the two reference-based methods. Reference-based methods are expected to learn fastest, since the demonstration places their starts on a known solution: on the \SI{0.4}{\meter} box (Figure~\ref{fig:bench_domains}(e)), Backplay converges fastest, and its advantage over RSI suggests that ordering the starts over training matters. BVER learns faster than RSI there and reaches 95\% success in roughly 65\% fewer training iterations than the best reference-free baseline. On the \SI{0.7}{\meter} box (Figure~\ref{fig:bench_domains}(f)), BVER is the only reference-free curriculum that learns the task, reaching about 90\% success after 4000 iterations, whereas the demonstration lets RSI and Backplay solve it in about 1600. Since this task requires two skills, walking and climbing, we hypothesize that the bidirectional expansion helps not only through spatial coverage but also by training both skills at the same time.

\paragraph{Robustness:}
We evaluate the policies trained on \textit{box climbing} under (1) initial yaw perturbations of up to $\pm 90^\circ$, (2) start positions sampled uniformly from the task space (goal set to the target goal), and (3) goal positions sampled uniformly from the task space (initialized on the initial distribution). Figure~\ref{fig:bench_domains}(g) shows that BVER's policy maintains a high success rate under all three variations.
RSI and Backplay train on states near the reference trajectory and thus see few states outside it, which likely explains their lower robustness.
RC varies only the start states and always trains on the target goal, so its lower goal robustness is expected. The sample-efficiency axis (time to 90\% success) is given relative to the fastest run over all methods and seeds (Backplay, 300 iterations).

\subsection{Training Coverage}\label{sec:exp_emergent}

Since BVER biases its expansion toward unexplored task space, we expect it to train the policy on the whole terrain rather than only on the easiest way to the goal. To test this, we design a task with three qualitatively different obstacles between the initial distribution and the goal: stairs in the center, pillars on one side, and a ramp on the other (Figure~\ref{fig:exp_renders_dtsg}).
The initial distribution spans the terrain's full width, so each obstacle is closest for roughly a third of the initial states.
Only on this task do we add a dense shaping term~\citep{ng1999policy} that rewards progress toward the goal, so that vanilla PPO finds a solution at all; both methods receive the same reward, and BVER's reward band still uses the sparse term only (Appendix~\ref{sec:appendix_obs_act}). We then roll out both policies from the initial distribution (one seed, 128 episodes each) and record which obstacle each successful episode traverses.
\begin{wrapfigure}[15]{r}{0.24\linewidth}
    \centering
    \vspace{-\baselineskip}
    \includegraphics[width=\linewidth]{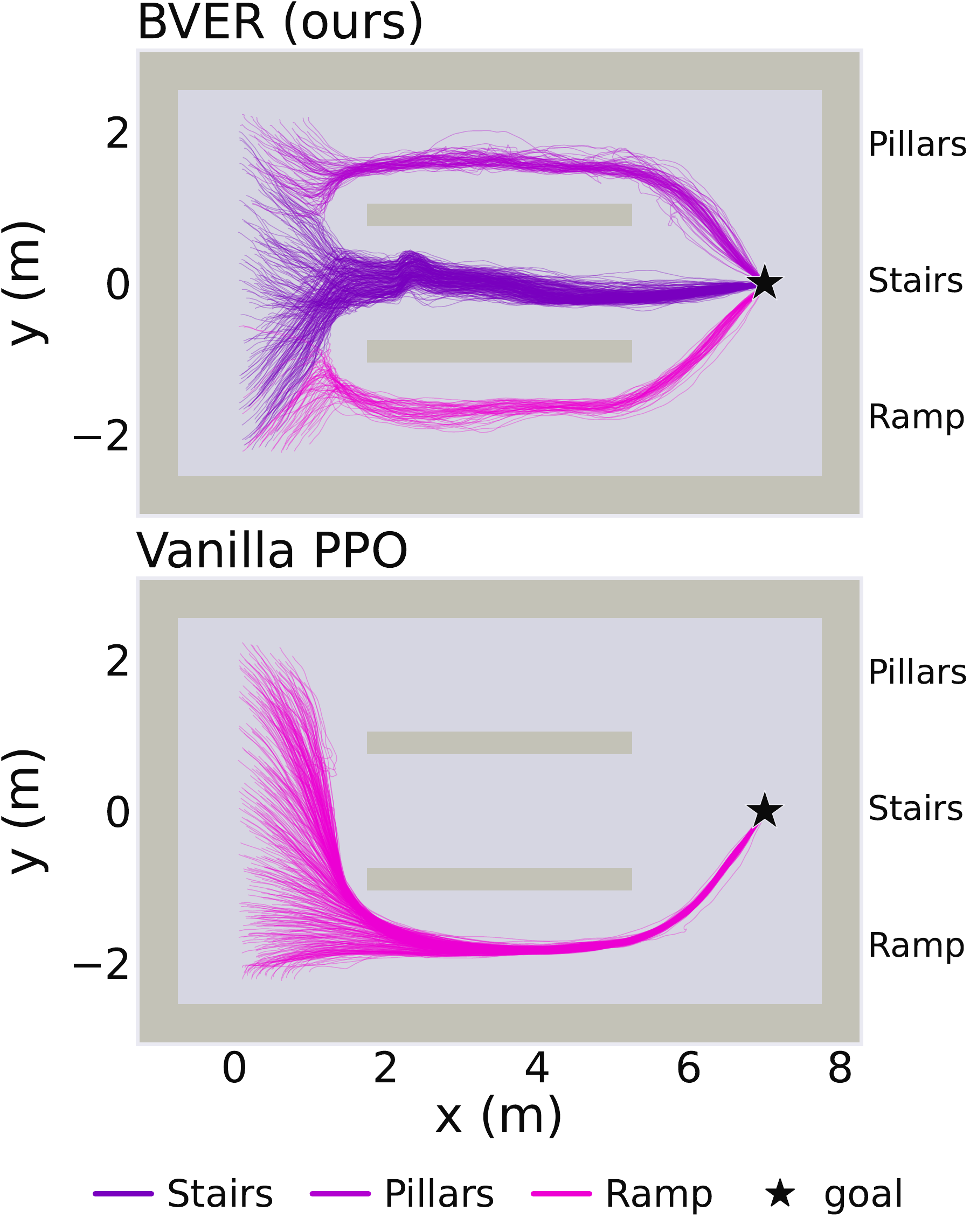}
    \caption{Successful trajectories of BVER and vanilla PPO on the \textit{three-route terrain}.}\label{fig:eval_path_diversity}
\end{wrapfigure}
Figure~\ref{fig:eval_path_diversity} shows that vanilla PPO only learns to traverse the ramp, the easiest of the three obstacles: regardless of the start, all successful episodes detour to it, even from in front of the stairs or pillars. In contrast, BVER's policy traverses all three obstacles (stairs 64\%, pillars 21\%, ramp 15\% of successful episodes), consistent with a curriculum that covers the whole terrain and trains the skills each obstacle requires.

\subsection{Scanned Terrains}\label{sec:exp_real_world}

\textit{Box climbing} involves a single, simple obstacle. To test whether its hyperparameters transfer directly to more complex geometry, we evaluate BVER on four terrains scanned from real environments (Figures~\ref{fig:exp_renders_freestanding_boulder_I}--\ref{fig:exp_renders_small_rocks}): two freestanding boulders to climb, a steep rocky slope, and a field of small stones, each requiring different locomotion skills. We reuse all PPO and BVER hyperparameters from \textit{box climbing} (Section~\ref{sec:exp_main}) without retuning; only the task space is chosen to roughly cover the terrain, the episode length (\SIrange{10}{12}{\second} instead of \SI{4}{\second}) and the number of training iterations are adapted to the larger terrains, and the number of environments is halved for training speed at the same exploit-to-explore ratio (Table~\ref{tab:bver_hyperparameters}). On the \textit{freestanding boulders} and the \textit{small stone field}, the initial distribution is chosen such that no path to the goal is favored. At the end of training (3000 iterations, 5000 on the \textit{rocky ascent}), BVER learns all four terrains, reaching a success rate (mean $\pm$ std.\ error over five seeds) of 91.0$\pm$1.7\% on \textit{freestanding boulder I}, 84.4$\pm$2.1\% on \textit{freestanding boulder II}, 77.2$\pm$2.2\% on the \textit{rocky ascent}, and 59.8$\pm$4.4\% on the \textit{small stone field}, which is the hardest of the four. Vanilla PPO, in contrast, does not succeed a single time on any of the four terrains, even with a higher entropy coefficient to encourage exploration. BVER's hyperparameters thus transfer across tasks within a domain.

\begin{figure}[t!]
    \centering
    \hfill
    \begin{subfigure}[t]{0.24\linewidth}
        \centering
        \includegraphics[width=\linewidth]{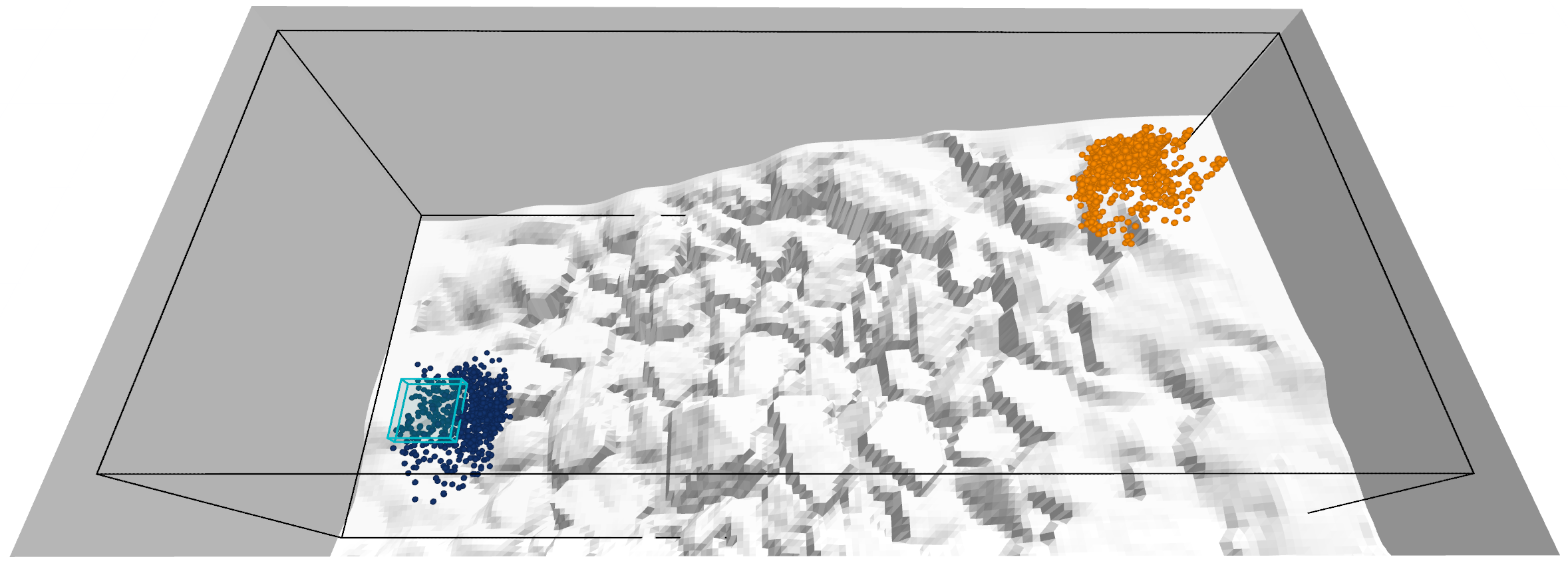}
        \caption{Iteration 150.}\label{fig:ra_150}
    \end{subfigure}
    \hfill
    \begin{subfigure}[t]{0.24\linewidth}
        \centering
        \includegraphics[width=\linewidth]{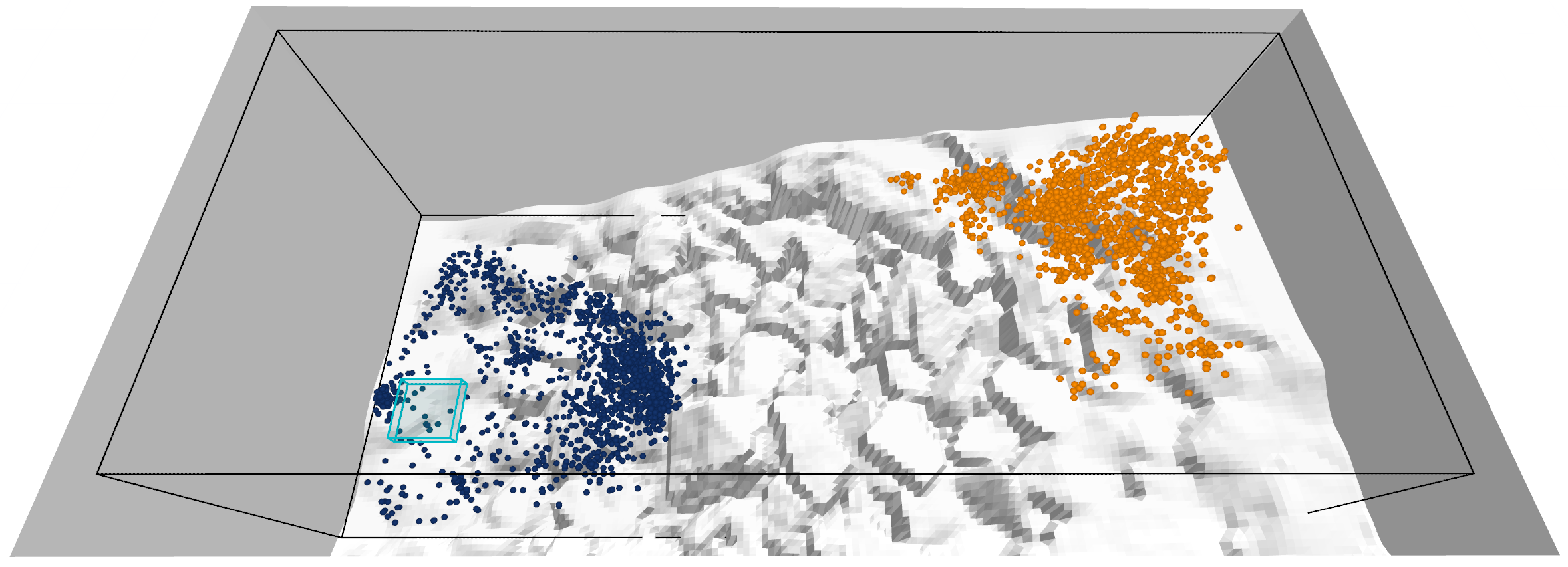}
        \caption{Iteration 500.}\label{fig:ra_500}
    \end{subfigure}
    \hfill
    \begin{subfigure}[t]{0.24\linewidth}
        \centering
        \includegraphics[width=\linewidth]{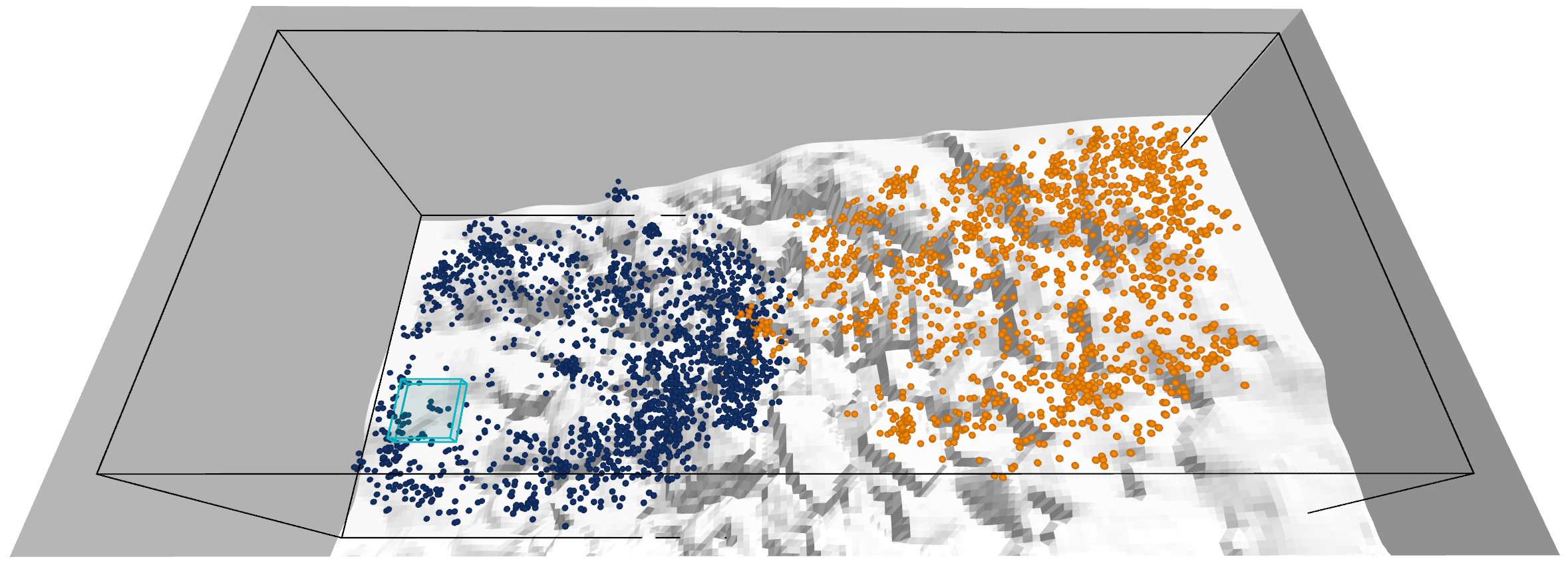}
        \caption{Iteration 1000.}\label{fig:ra_1000}
    \end{subfigure}
    \hfill
    \begin{subfigure}[t]{0.24\linewidth}
        \centering
        \includegraphics[width=\linewidth]{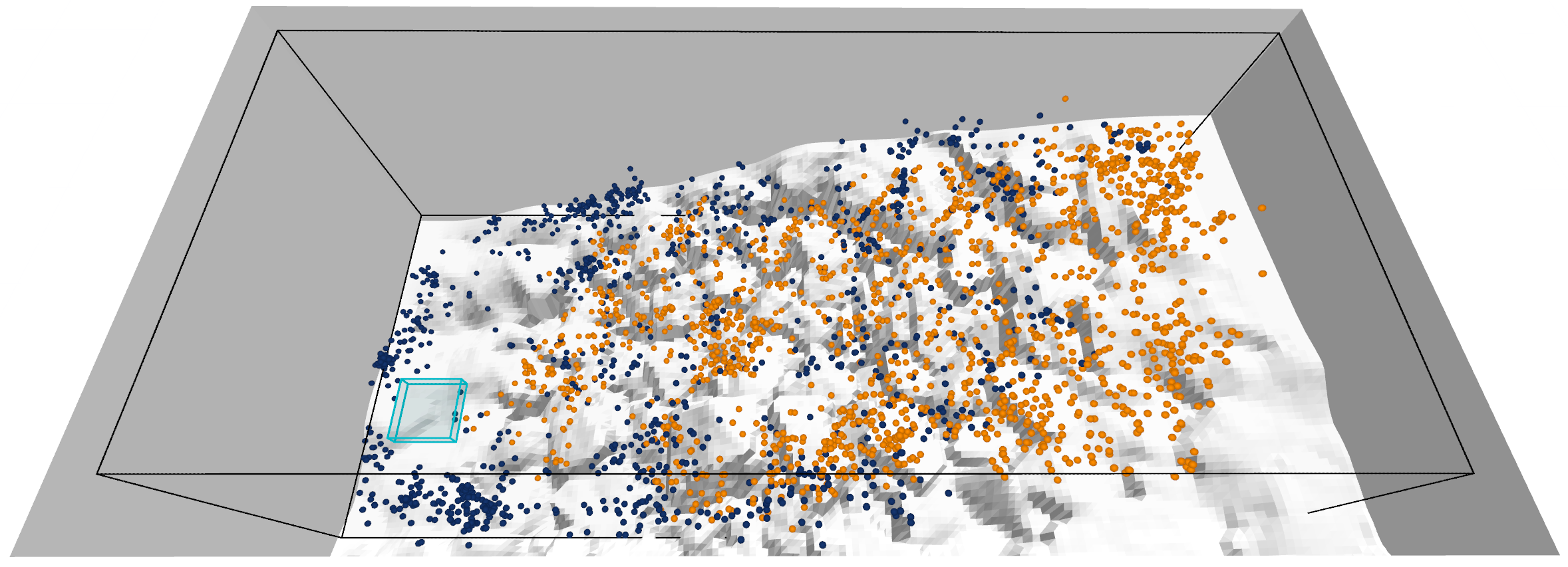}
        \caption{Iteration 2000.}\label{fig:ra_2000}
    \end{subfigure}
    \hfill\strut{}
    \caption{Samples of the explored regions on the \textit{rocky ascent} terrain at different training iterations. Blue indicates the intermediate goals (forwards expansion), orange the intermediate starts (backwards expansion), cyan the initial distribution, and the black box represents the task space.}\label{fig:scans}
\end{figure}

Figure~\ref{fig:scans} shows the explored regions on the \textit{rocky ascent} terrain over training. The intermediate starts and goals begin clustered around their roots and gradually spread over the reachable terrain as the policy improves. Although the task space (black box) is much larger than the reachable area, the random walks find a path to the goal without prior knowledge of the terrain, and the starts and goals cover the terrain rather than a single path, so the curriculum trains the policy on the whole terrain rather than only on the target task.

\subsection{Ablation Studies}\label{sec:exp_ablations}

To isolate the contribution of BVER's components, we ablate the expansion direction on the three \textit{point-mass mazes}, \textit{ring-on-peg transfer}, and \textit{box climbing}, and the connect step on the mazes.

\begin{table}[!ht]
    \centering
    \caption{Ablation study: iterations until the mean success rate over five seeds reaches 75\% ($\infty$: not reached; n/a: not evaluated). PM-U/N/L: U-, N-, and large point-mass maze; RPT: ring-on-peg transfer; CU/CD: climbing up/down a box of the given height.}\label{tab:ablation_table}
    \setlength{\tabcolsep}{4pt}
    \begin{tabularx}{\linewidth}{@{}p{0.2\linewidth}cccccccc@{}}
        \toprule
                          & \multicolumn{8}{c}{Iterations to 75\% Success Rate ($\downarrow$)}                                                                                                           \\
                          & PM-U                                                               & PM-N         & PM-L         & RPT          & CU 0.4 m     & CU 0.7 m      & CD 0.4 m     & CD 0.7 m     \\
        \cmidrule(lr){2-9}
        BVER (ours)       & \textbf{160}                                                       & \textbf{800} & \textbf{140} & \textbf{120} & \textbf{600} & \textbf{3800} & \textbf{300} & \textbf{400} \\
        \cmidrule(lr){1-9}
        BVER fwd.\ only   & 380                                                                & 1100         & 440          & $\infty$     & 800          & $\infty$      & 400          & 600          \\
        BVER bwd.\ only   & 440                                                                & $\infty$     & 400          & $\infty$     & 800          & \textbf{3800} & 1700         & 2800         \\
        \cmidrule(lr){1-9}
        BVER (no connect) & 340                                                                & 1000         & 200          & n/a          & n/a          & n/a           & n/a          & n/a          \\
        BVER (greedy)     & 260                                                                & 1500         & 240          & n/a          & n/a          & n/a           & n/a          & n/a          \\
        \bottomrule
    \end{tabularx}
\end{table}

\paragraph{Expansion direction:}
We compare BVER against forwards and backwards expansion alone (fwd.\ and bwd.\ only), each with the full environment budget, including \textit{box climbing} up and down at \SI{0.4}{\meter} and \SI{0.7}{\meter}. The bidirectional expansion reaches 75\% success first on every task, tied only with the backwards expansion on climbing up the \SI{0.7}{\meter} box, whereas each direction alone fails to reach 75\% success on two of the eight tasks (Table~\ref{tab:ablation_table}). Which direction works depends on the task. On \textit{box climbing}, the expansion that starts on top of the box, i.e., backwards when climbing up and forwards when climbing down, is clearly stronger, likely because the random walks are subject to gravity and thus traverse the box edge more easily downwards than upwards: at \SI{0.7}{\meter}, the expansion from the ground never reaches 75\% success when climbing up and needs seven times more iterations than BVER when climbing down. On \textit{ring-on-peg transfer}, neither direction alone reaches 75\% success, as each expands from its peg into the free space in between but rarely reaches the other peg. In the mazes, BVER needs 1.4 to 2.9 times fewer iterations than the better unidirectional variant, and the backwards expansion alone fails in the \textit{N-maze}. In the \textit{large maze}, we observe that the unidirectional variants cover the space uniformly, whereas BVER focuses on the region between start and goal. Combining both directions thus removes the need to know in advance which direction suits a task. The learning curves are given in Appendix~\ref{sec:appendix_ablations}.

\paragraph{Connecting the frontiers:}
On the three mazes, we compare the default connect ratio $p_{\text{connect}} = 0.5$ against seeding the random walks only from uniform task space samples ($p_{\text{connect}} = 0$, no connect) or only from the other direction's solved set ($p_{\text{connect}} = 1$, greedy). Both extremes slow down learning on all three mazes (Table~\ref{tab:ablation_table}). Without connecting, the two frontiers spread evenly over $\mathcal{T}$ and likely take longer to meet, requiring up to 2.1 times more iterations. Greedy connecting drops the coverage bias and requires up to 1.9 times more iterations, most notably in the \textit{N-maze}, likely because its walls block the direct way toward the other frontier. Both coverage and connection thus contribute, and we use $p_{\text{connect}} = 0.5$ in all experiments (Table~\ref{tab:bver_hyperparameters}).
\section{Conclusion}\label{sec:conclusion}

We propose BVER, a curriculum for sparse-reward, long-horizon goal-conditioned reinforcement learning that transfers bidirectional expansion from sampling-based planning~\citep{lavalle2001randomized, kuffner2000rrt} to curriculum generation: it grows start states from the goal and goals from the initial state distribution, biases both toward unexplored task space, and steers them toward each other. In simulation, this yields sample-efficient and robust policies in navigation, locomotion, and manipulation.
The random walks run in the same parallelized simulator as the policy rollouts and increase the wall-clock training time by about 28\% in our experiments.

BVER relies on a simulator that can be reset to arbitrary states, on random-action rollouts, and on a bounded task space with a meaningful distance metric. All task spaces in this work are two- or three-dimensional position boxes with a Euclidean metric; higher-dimensional or non-Euclidean task spaces remain untested and could use learned distances such as the dynamical distance of~\citet{hartikainen2020dynamicaldistancelearningsemisupervised}. Like RC~\citep{florensa2017reverse}, BVER's backwards expansion further assumes approximately reversible dynamics: a random walk from the goal reaching a state does not imply that the goal is reachable from that state, as the gravity asymmetry in \textit{box climbing} illustrates, and such states are dropped only after a rollout has shown them unsolvable ($R \le R_{\min}$); learned dynamics models or rollouts of the current policy could replace the random walks. Finally, all results are in simulation, and hardware deployment is left to future work.

\subsection*{AI use statement}
In this work, we used generative AI tools
for creating scientific figures from non-AI-generated data, creating and
editing software code, and editing the paper text to improve readability
(proofreading of grammar and style). We have reviewed all AI-assisted work:
generated figures were checked against the underlying data, AI-assisted code
was verified and tested for correctness by the authors, and all suggested text
edits were reviewed and accepted by the authors.

\subsection*{Reproducibility statement}
Experiments can be reproduced with the code provided in the project website. The code includes all hyperparameters, random seeds, and instructions to run the experiments. We run all experiments on a single NVIDIA RTX 4090 GPU with five different random seeds each.

\bibliography{references}
\bibliographystyle{preprint}

\appendix
\section{Appendix}\label{sec:appendix}

\subsection{Algorithms}\label{sec:appendix_algorithms}

Algorithms~\ref{alg:random_walk}--\ref{alg:mix} give the subroutines of the training loop in
Algorithm~\ref{alg:bver_overview}, each with the hyperparameters it uses
(Table~\ref{tab:bver_hyperparameters}).
Each direction keeps the same four sets of states: a candidate set $C$ of at most $M_C$
random-walk states, a frontier buffer $F$ holding the
states whose rollout return lies strictly between $R_{\min}$ and $R_{\max}$, a solved buffer $S$
holding the states with a return of at least $R_{\max}$, and the mixed set $\mathcal{I}$ from
which the exploitation rollouts are drawn. The superscripts $\text{fwd}$ and
$\text{bwd}$ distinguish the forwards from the backwards expansion; states in the
forwards buffers enter the rollouts as intermediate goals through $\phi$, states in the
backwards buffers as intermediate start states.
Algorithm~\ref{alg:random_walk} expands a frontier by simulating an action-noise random
walk from the nearest neighbor of a sampled task space point,
Algorithm~\ref{alg:threshold} sorts the collected rollouts into $F$ and $S$ by their
return, and Algorithm~\ref{alg:mix} rebuilds $\mathcal{I}$ from the two buffers,
replaying a fraction $\nu$ of solved states.

Three details are worth noting. First, in the forwards direction, \textsc{Threshold} stores the
final state of a rollout rather than its commanded goal: the forwards buffers seed random walks,
which must be reset to full simulator states, whereas a commanded goal is only a point in
$\mathcal{G}$. The return of the commanded goal decides whether the rollout is sorted into $F$
or $S$, so the stored state is the state the policy actually reached while attempting a goal of
the corresponding difficulty, a form of hindsight relabeling; for a solved rollout, which reached
and held its goal, this state typically lies within $\epsilon$ of the commanded goal. Second, the
connect samples in \textsc{RandomWalk} are drawn from the other direction's solved buffer
$S^{\bar{d}}$, i.e., from states the policy already reaches from $\rho_0$ or from which it already
reaches $g^\star$. Third, exploitation is optional: with $N_{\text{exploit}} = 0$,
the exploitation rollouts are omitted and the mixed set serves only to seed the random walks.
We use this setting on the mazes and \textit{ring-on-peg transfer}, where exploration rollouts alone did
not lead to forgetting (Table~\ref{tab:bver_hyperparameters}).

\begin{algorithm}[H]
    \caption{Random walk frontier expansion}\label{alg:random_walk}
    \small
    \begin{algorithmic}[1]
        \Hyperparameters{} connect ratio $p_{\text{connect}}$, action noise covariance $\Sigma$, random walk horizon $T_B$, number of subsamples per walk $N_s$
        \Procedure{RandomWalk}{$\mathcal{I}, \mathcal{T}, S^{\bar{d}}$}
        \If{$u < p_{\text{connect}}$, \quad $u \sim \mathcal{U}[0,1]$}
        \State{} $x_{\text{rand}} \gets \phi(\Call{Sample}{S^{\bar{d}}})$ \Comment{connect: sample from the other direction's solved buffer}
        \Else{}
        \State{} $x_{\text{rand}} \gets \Call{Sample}{\mathcal{T}}$ \Comment{sample uniformly from task space}
        \EndIf{}
        \State{} $s_0 \gets \Call{NearestNeighbor}{x_{\text{rand}}, \mathcal{I}}$ \Comment{w.r.t.\ $\lVert \phi(\cdot) - x_{\text{rand}} \rVert$}
        \For{$t = 1, \ldots{}, T_B$}
        \State{} $a_t \sim \mathcal{N}(0, \Sigma)$ \Comment{random action}
        \State{} $s_t \sim P(s_t \mid s_{t-1}, a_t)$ \Comment{simulate random walk from nearest neighbor}
        \EndFor{}
        \State{} $\mathcal{C} \gets \Call{Subsample}{\{s_1, \ldots{}, s_{T_B}\}, N_s}$ \Comment{new candidates: $N_s$ subsampled states}
        \State{} \Return{} $\mathcal{C}$
        \EndProcedure{}
    \end{algorithmic}
\end{algorithm}

\begin{algorithm}[H]
    \caption{Return-based thresholding of rollouts into frontier and solved buffers}\label{alg:threshold}
    \small
    \begin{algorithmic}[1]
        \Hyperparameters{} return band $R_{\min}, R_{\max}$, maximum buffer sizes $M_F, M_S$
        \Procedure{Threshold}{$D, F, S, d$}
        \For{$(\tau, g) \in D$} \Comment{explore and exploit rollouts with return $R(\tau, g)$}
        \If{$d = \text{fwd}$}
        \State{} $x \gets \tau.\Call{Last}{\,}$ \Comment{final state, i.e., the achieved goal (Appendix~\ref{sec:appendix_algorithms})}
        \Else{}
        \State{} $x \gets \tau.\Call{First}{\,}$ \Comment{start state}
        \EndIf{}
        \If{$R_{\min} < R(\tau, g) < R_{\max}$}
        \State{} $F.\Call{Append}{x}$ \Comment{frontier, at most $M_F$ states}
        \ElsIf{$R(\tau, g) \geq R_{\max}$}
        \State{} $S.\Call{Append}{x}$ \Comment{solved, at most $M_S$ states}
        \EndIf{}
        \EndFor{}
        \State{} \Return{} $F, S$
        \EndProcedure{}
    \end{algorithmic}
\end{algorithm}

\begin{algorithm}[H]
\caption{Rebuild of the mixed set from frontier and solved buffers}\label{alg:mix}
\small
\begin{algorithmic}[1]
\Hyperparameters{} replay ratio $\nu$, mixed set size $M_I$
\Procedure{Mix}{$F, S$}
    \State{} $n_S \gets \min\left(\lfloor \nu \, M_I \rfloor, \; |S|\right)$ \Comment{replayed solved states}
    \State{} $n_F \gets \min\left(M_I - n_S, \; |F|\right)$ \Comment{rest from the frontier}
    \State{} $\mathcal{I} \gets \Call{Subsample}{S, n_S} \cup \Call{Subsample}{F, n_F}$ \Comment{uniform, without replacement}
    \State{} \Return{} $\mathcal{I}$
\EndProcedure{}
\end{algorithmic}
\end{algorithm}

\begin{figure}[t]
    \centering
    \includegraphics[width=\linewidth]{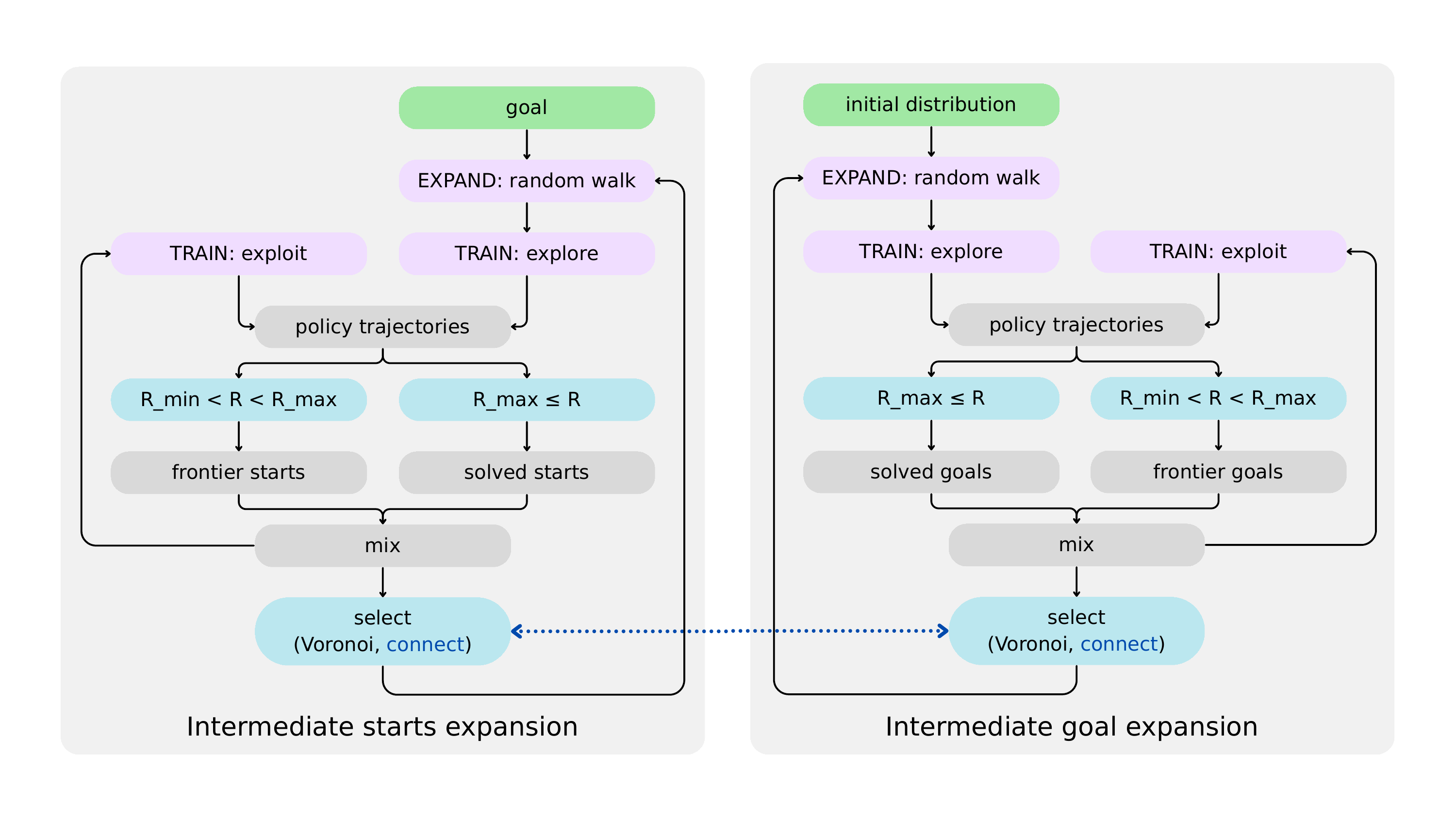}
    \caption{Overview of BVER. The backwards expansion (left) grows intermediate starts from the
        goal, the forwards expansion (right) intermediate goals from the initial distribution. In
        each, random walks expand the selected seeds, the policy trains on the new states (explore)
        and on the mixed set (exploit), the rollouts are sorted by return into frontier and solved
        sets, and the two are mixed. The next seeds are selected by the Voronoi bias or, with
        probability $p_{\text{connect}}$, toward the other expansion's solved set (dotted arrow).}\label{fig:method}
\end{figure}

\subsection{Experiment setup}\label{sec:appendix_obs_act}
All policies are trained with the PPO implementation of the RSL-RL framework~\citep{schwarke2025rslrllearninglibraryrobotics}. Table~\ref{tab:ppo_hyperparameters} lists the PPO hyperparameters and Table~\ref{tab:bver_hyperparameters} the BVER hyperparameters of all experiments, using these abbreviations: Point-Maze (PM), Climb Up (CU), Climb Down (CD), Three-Route Terrain (TRT), Freestanding Boulder I (FB I), Freestanding Boulder II (FB II), Rocky Ascent (RA), Small Stone Field (SSF), and Ring-on-peg Transfer (RPT).

\paragraph{Point-Maze (PM).}
The agent is a velocity-controlled point mass in a walled grid with \SI{4}{\meter} cells, controlled at \SI{20}{\hertz}. The observation is six-dimensional and noise-free: the planar position, the planar velocity, and the vector to the goal; actor and critic receive the same input. The action is a two-dimensional planar velocity command. The reward consists of a single sparse term that is nonzero only within \SI{1}{\meter} of the goal, where it decreases linearly from 1 at the goal to 0 at this tolerance. The episode horizon is \SI{9}{\second} in the \textit{U-maze}, \SI{32}{\second} in the \textit{N-maze}, and \SI{15}{\second} in the \textit{large maze}.

\paragraph{Climb Up/Down (CU, CD) and scans (FB I, FB II, RA, SSF).}
These tasks use ANYmal D with a \SI{50}{\hertz} joint-position controller; the 12-dimensional action is an offset from the default joint configuration. The policy observes the base position and angle, the base linear and angular velocity, the projected gravity, the three-dimensional goal position in the base frame, the joint positions and velocities, the previous action, and a height scan over a \SI{2}{\meter}~$\times$~\SI{1}{\meter} grid at \SI{0.1}{\meter} resolution. The policy observations are perturbed with uniform noise; the critic receives the same terms without noise. The reward is sparse: a goal term that decreases linearly from 1 at the goal to 0 at a distance of \SI{0.25}{\meter} and is zero beyond, and a termination penalty for illegal body contact or a bad base orientation. No shaping terms are used. Episodes last \SI{4}{\second} on the climb tasks, \SI{12}{\second} on RA, and \SI{10}{\second} on the other scans.

\paragraph{Three-Route Terrain (TRT).}
TRT shares the robot, the action space, and the observations, including the base position and yaw, with the tasks above. Unlike the climb and scan tasks, the reward is not purely sparse. In addition to the same sparse goal term, a shaping term based on the distance potential $-\lVert \phi(s) - g \rVert$~\citep{ng1999policy} is added to provide a reward signal strong enough for the no-curriculum PPO baseline to find solutions; vanilla PPO and BVER receive the same reward. The term rewards the closing speed, i.e., the base velocity $v$ toward the goal, clipped to \SI{1.5}{\meter\per\second} and set to zero within \SI{0.25}{\meter} of the goal:
\begin{equation*}
    r_{\mathrm{progress}}(s, g) = \begin{cases} \operatorname{clip}\!\left( v^{\top} \frac{g - \phi(s)}{\lVert g - \phi(s) \rVert}, -1.5, 1.5 \right) & \text{if } \lVert g - \phi(s) \rVert > \SI{0.25}{\meter}, \\ 0 & \text{otherwise.} \end{cases}
\end{equation*}
BVER's reward band is evaluated on the sparse term only. Episodes last \SI{10}{\second}. For the route analysis in Section~\ref{sec:exp_emergent}, we roll out each policy after 2000 training iterations (one seed) in 128 episodes and assign a successful episode to the stairs, pillars, or ramp if the base passes through a coordinate box around the respective obstacle.

\paragraph{Ring-on-peg Transfer (RPT).}
A Franka Panda with stiff joints holds a ring that is rigidly attached to its hand and must move it from the bottom of one peg to the bottom of another; control runs at \SI{30}{\hertz}. The 27-dimensional observation consists of the seven arm joint positions and velocities, the ring center, the vector from the ring center to the goal, and the previous action. The seven-dimensional action is a relative joint-position target, so a zero action holds the current pose. The reward is the same as in the \textit{point-mass mazes}, measured in three dimensions on the ring center with a tolerance of \SI{3}{\centi\meter}. The episode horizon is \SI{8}{\second}.

\paragraph{Baselines.}
All baselines use the PPO hyperparameters of Table~\ref{tab:ppo_hyperparameters} and the same number of environments and simulation steps per iteration as BVER. The random curriculum draws intermediate starts and goals from a collection of states sampled offline uniformly from the task space and kept if they are collision-free; it does not filter them by return. RSI and Backplay reset to states of a single demonstration, a successful rollout of a trained BVER policy; any successful policy could provide it. We implemented RC in Isaac Lab following the original algorithm~\citep{florensa2017reverse} as closely as possible.

\begin{table}[!ht]
    \centering
    \caption{PPO hyperparameters for all experiments.}\label{tab:ppo_hyperparameters}
    \renewcommand{\arraystretch}{1.3}
    \begin{tabularx}{\linewidth}{@{}p{0.25\linewidth}c|c|c@{}}
        \toprule
                                 & \multicolumn{3}{c}{Experiments}                                                        \\
        \cmidrule(lr){2-4}
        Hyperparameters          & PM                              & CU, CD, TRT, FB I, FB II, RA, SSF & RPT              \\
        \cmidrule(lr){1-4}
        MLP hidden dimensions    & $[128, 64, 16]$                 & $[512, 256, 128]$                 & $[256, 128, 64]$ \\
        Activation function      & \textit{ELU}                    & \textit{ELU}                      & \textit{ELU}     \\
        Value loss coefficient   & $1.0$                           & $1.0$                             & $1.0$            \\
        Clip parameter           & $0.2$                           & $0.2$                             & $0.2$            \\
        Entropy coefficient      & $0.01$                          & $0.0025$                          & $0.002$          \\
        Learning rate            & $5\cdot 10^{-4}$                & $5\cdot 10^{-4}$                  & $5\cdot 10^{-4}$ \\
        Discount factor $\gamma$ & $0.99$                          & $0.99$                            & $0.99$           \\
        $\lambda$ (GAE)          & $0.95$                          & $0.95$                            & $0.95$           \\
        Desired KL divergence    & $0.01$                          & $0.01$                            & $0.01$           \\
        \bottomrule
    \end{tabularx}
    \renewcommand{\arraystretch}{1}
\end{table}

\begin{table}[!ht]
    \centering
    \caption{BVER hyperparameters for all experiments. $R_{\min}$ and $R_{\max}$ are thresholds on the normalized return of Equation~\ref{eq:normalized_return}. The connect ratio $p_{\text{connect}}$ and the number of subsamples per walk $N_s$ are shared across all tasks; the other hyperparameters are set per domain, and $T_B$ and the environment split also differ between the mazes. All hyperparameters were selected empirically per domain in preliminary runs.}\label{tab:bver_hyperparameters}
    \setlength{\tabcolsep}{3pt}
    \begin{tabularx}{\linewidth}{@{}p{0.26\linewidth}ccc|c|cc|c@{}}
        \toprule
                                         & \multicolumn{7}{c}{Experiments}                                                                                                                 \\
        \cmidrule(lr){2-8}
        Hyperparameters                  & PM-U                            & PM-N            & PM-L            & TRT             & CU, CD          & FB I, FB II, RA, SSF & RPT            \\
        \cmidrule(lr){1-8}
        Env split $N_{\mathrm{exploit}}$ & $0$                             & $0$             & $0$             & $2048$          & $2048$          & $1024$               & $0$            \\
        Env split $N_{\mathrm{explore}}$ & $512$                           & $4096$          & $4096$          & $4096$          & $4096$          & $2048$               & $2048$         \\
        $R_{\min}$                       & $0.02$                          & $0.02$          & $0.02$          & $0.05$          & $0.05$          & $0.05$               & $0.1$          \\
        $R_{\max}$                       & $0.95$                          & $0.95$          & $0.95$          & $0.2$           & $0.2$           & $0.2$                & $0.85$         \\
        Replay ratio $\nu$               & $0.2$                           & $0.2$           & $0.2$           & $0.05$          & $0.05$          & $0.05$               & $0.1$          \\
        $p_{\mathrm{connect}}$           & $0.5$                           & $0.5$           & $0.5$           & $0.5$           & $0.5$           & $0.5$                & $0.5$          \\
        Frontier buffer size $M_F$       & $10^3$                          & $10^3$          & $10^3$          & $2\cdot10^3$    & $2\cdot10^3$    & $2\cdot10^3$         & $10^3$         \\
        Solved buffer size $M_S$         & $10^3$                          & $10^3$          & $10^3$          & $10^4$          & $10^4$          & $10^4$               & $10^3$         \\
        Candidate set size $M_C$         & $500$                           & $500$           & $500$           & $5\cdot10^3$    & $5\cdot10^3$    & $5\cdot10^3$         & $500$          \\
        Mixed set size $M_I$             & $10^3$                          & $10^3$          & $10^3$          & $2\cdot10^3$    & $2\cdot10^3$    & $2\cdot10^3$         & $10^3$         \\
        Subsamples per walk $N_s$        & $4$                             & $4$             & $4$             & $4$             & $4$             & $4$                  & $4$            \\
        Random walk horizon $T_B$        & $80$                            & $200$           & $80$            & $128$           & $64$            & $64$                 & $10$           \\
        Action noise $\Sigma$            & $10\cdot\bf{I}$                 & $10\cdot\bf{I}$ & $10\cdot\bf{I}$ & $10\cdot\bf{I}$ & $10\cdot\bf{I}$ & $10\cdot\bf{I}$      & $3\cdot\bf{I}$ \\
        \bottomrule
    \end{tabularx}
\end{table}

\subsection{Ablation learning curves}\label{sec:appendix_ablations}

\begin{figure}[!ht]
    \centering
    \includegraphics[width=\linewidth]{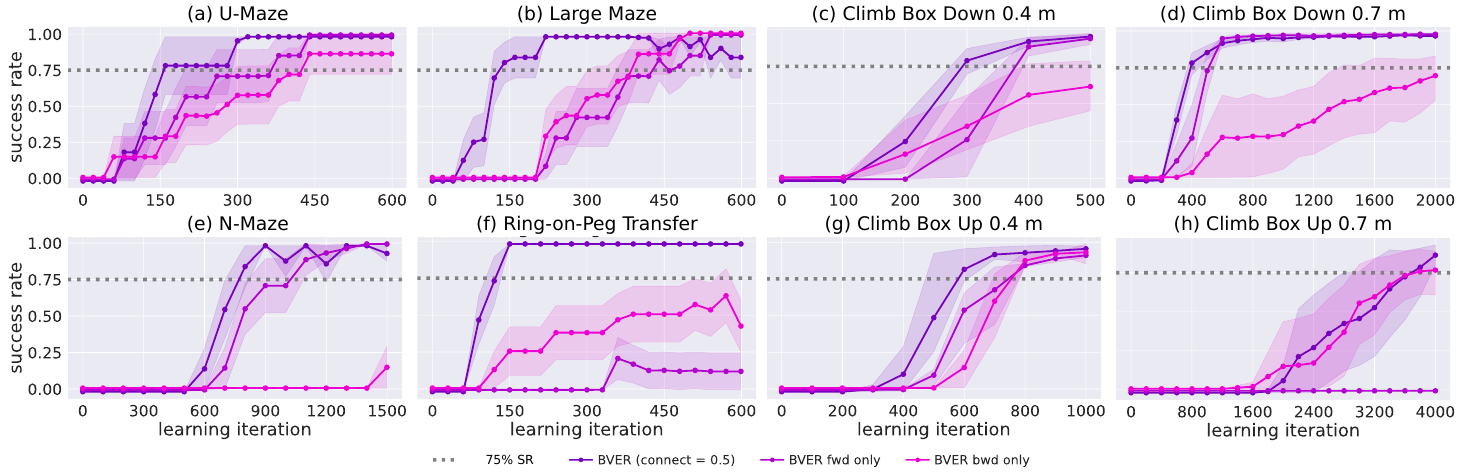}
    \caption{Expansion direction ablation. Combining forwards and backwards expansions reaches 75\% success first on all tasks, tied with the backwards expansion on climbing up the \SI{0.7}{\meter} box (h). Which direction alone works depends on the task: on \textit{box climbing} (c, d, g, h), the expansion starting on top of the box is stronger, and on \textit{ring-on-peg transfer} (f), neither direction alone reaches 75\% success. The dashed line marks the 75\% success rate (SR) used in Table~\ref{tab:ablation_table}.}\label{fig:eval_ablations}
\end{figure}

\begin{figure}[!ht]
    \centering
    \hfill
    \includegraphics[width=0.99\linewidth]{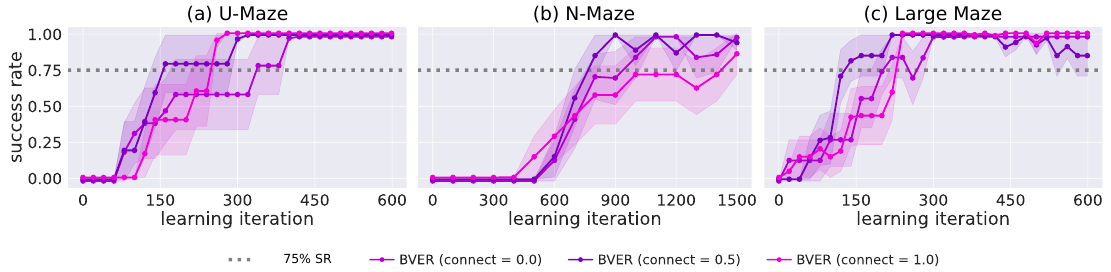}
    \hfill
    \caption{Connect ratio ablation on the three mazes. The default $p_{\text{connect}} = 0.5$ learns fastest on all three mazes; seeding the random walks only from uniform task space samples ($p_{\text{connect}} = 0$) or only from the other direction's solved set ($p_{\text{connect}} = 1$) slows down learning. The dashed line marks the 75\% success rate (SR) used in Table~\ref{tab:ablation_table}.}\label{fig:maze_ablation_connect}
\end{figure}

Figures~\ref{fig:eval_ablations} and~\ref{fig:maze_ablation_connect} show the learning curves
of the ablations summarized in Table~\ref{tab:ablation_table}. With a single expansion
direction, the success rate often plateaus below that of BVER, most clearly on \textit{ring-on-peg
transfer}, where each unidirectional variant expands from its peg into the free space in
between but rarely reaches the other peg. For the connect step, both $p_{\text{connect}} = 0$
and $p_{\text{connect}} = 1$ rise later than the default $p_{\text{connect}} = 0.5$ on all
three mazes: without connecting, the two frontiers spread evenly over $\mathcal{T}$ and likely take
longer to meet, and with greedy connecting, the expansion loses its coverage bias.

\end{document}